\documentclass[journal]{IEEEtran}
\usepackage{fix-cm}
\usepackage{url}
\usepackage[hidelinks,hypertexnames=false]{hyperref}
\usepackage{graphicx}
\usepackage{cite}
\usepackage{booktabs}
\usepackage{multirow}
\usepackage{amssymb}
\usepackage{makecell}
\usepackage{enumitem}
\usepackage{ulem}
\usepackage{color}
\usepackage{float}
\usepackage{algorithm}
\usepackage{cuted}
\usepackage{capt-of}
\usepackage{scalerel}
\usepackage{tikz}
\usetikzlibrary{svg.path}
\definecolor{orcidlogocol}{HTML}{A6CE39}
\tikzset{
  orcidlogo/.pic={
    \fill[orcidlogocol] svg{M256,128c0,70.7-57.3,128-128,128C57.3,256,0,198.7,0,128C0,57.3,57.3,0,128,0C198.7,0,256,57.3,256,128z};
    \fill[white] svg{M86.3,186.2H70.9V79.1h15.4v48.4V186.2z}
                 svg{M108.9,79.1h41.6c39.6,0,57,28.3,57,53.6c0,27.5-21.5,53.6-56.8,53.6h-41.8V79.1z M124.3,172.4h24.5c34.9,0,42.9-26.5,42.9-39.7c0-21.5-13.7-39.7-43.7-39.7h-23.7V172.4z}
                 svg{M88.7,56.8c0,5.5-4.5,10.1-10.1,10.1c-5.6,0-10.1-4.6-10.1-10.1c0-5.6,4.5-10.1,10.1-10.1C84.2,46.7,88.7,51.3,88.7,56.8z};
  }
}
\newcommand\orcidicon[1]{\href{https://orcid.org/#1}{\mbox{\scalerel*{
\begin{tikzpicture}[yscale=-1,transform shape]
\pic{orcidlogo};
\end{tikzpicture}
}{|}}}}
\usepackage{hyperref}
\hypersetup{
    colorlinks=true,
    linkcolor=blue,
    filecolor=black,      
    urlcolor=black,
    citecolor=blue
}
\usepackage{tikz-network}
\newcolumntype{L}[1]{>{\raggedright\arraybackslash}p{#1}}
\newcolumntype{C}[1]{>{\centering\arraybackslash}p{#1}}
\newcolumntype{R}[1]{>{\raggedleft\arraybackslash}p{#1}}

\newcommand{\settablefont}{\fontsize{6.9}{11.8}\selectfont}
\newcommand{\egi}{\textit{e.g.}}
\newcommand{\iei}{\textit{i.e.}}

\newcommand{\etal}{\textit{et al.}}

\begin{document}
\title{Environment-Driven Online LiDAR-Camera Extrinsic Calibration}
\normalem
\author{
Zhiwei Huang$^{\orcidicon{0009-0008-7084-052X}\,}$,
Jiaqi Li$^{\orcidicon{0009-0008-5496-8136}\,}$, 
Hongbo Zhao$^{\orcidicon{0009-0008-2198-2484}\,}$,
Xiao Ma$^{\orcidicon{0009-0000-1058-2014}\,}$,\\
Ping Zhong$^{\orcidicon{0000-0003-3393-8874}\,}$,
Xiaohu Zhou$^{\orcidicon{0000-0002-7602-4848}\,}$,
Wei Ye$^{\orcidicon{0000-0002-3784-7788}\,}$,
and Rui Fan$^{\orcidicon{0000-0003-2593-6596}\,}$,~\IEEEmembership{Senior Member,~IEEE}
\thanks{This research was supported by the National Natural Science Foundation of China under Grants 62473288, 62233013, 62272489, and 62388101, National Key Research and Development Program of China (2025YFE0200003), the Fundamental Research Funds for the Central Universities, Xiaomi Young Talents Program, and the National Key Laboratory of Human-Machine Hybrid Augmented Intelligence, Xi'an Jiaotong University under Grant No. HMHAI-202406. (\emph{Corresponding author: Rui Fan})}
\thanks{Zhiwei Huang, Jiaqi Li, Hongbo Zhao, and {Wei Ye} are with the Department of Control Science \& Engineering, the College of Electronic \& Information Engineering, Tongji University, Shanghai 201804, China (e-mails: \{2431985, 2251550, {hongbozhao}, {yew}\}@tongji.edu.cn).}
\thanks{Ping Zhong is with the School of Computer Science and Engineering, Central South University, Changsha 410083, Hunan, China, as well as with the National Key Laboratory of Science and Technology on Automatic Target Recognition, National University of Defense Technology, Changsha 410073, Hunan, China (e-mail: ping.zhong@csu.edu.cn).}
\thanks{Xiao Ma is with the Beijing Institute of Aerospace Control Devices, Beijing 100039, China (e-mail: mx\_169@126.com).}
\thanks{Xiaohu Zhou is with the Institute of Automation, Chinese Academy of Sciences, Beijing 100190, China (e-mail: xiaohu.zhou@ia.ac.cn).}
\thanks{Rui Fan is with the College of Electronic and Information Engineering, Shanghai Institute of Intelligent Science and Technology, Shanghai Research Institute for Intelligent Autonomous Systems, Shanghai Key Laboratory of Intelligent Autonomous Systems, State Key Laboratory of Autonomous Intelligent Unmanned Systems, and Frontiers Science Center for Intelligent Autonomous Systems (Ministry of Education), Tongji University, Shanghai 201804, China, as well as with the National Key Laboratory of Human-Machine Hybrid Augmented Intelligence, Xi'an Jiaotong University, Xi'An, Shaanxi 710049, China (e-mail: rui.fan@ieee.org).
}
}
\maketitle

\begin{abstract}
LiDAR-camera extrinsic calibration (LCEC) is crucial for multi-modal data fusion in autonomous robotic systems. Existing methods, whether target-based or target-free, typically rely on customized calibration targets or fixed scene types, which limit their applicability in real-world scenarios.
To address these challenges, we present EdO-LCEC, the first environment-driven online calibration approach. Unlike traditional target-free methods, EdO-LCEC employs a generalizable scene discriminator to estimate the feature density of the application environment. Guided by this feature density, EdO-LCEC extracts LiDAR intensity and depth features from varying perspectives to achieve higher calibration accuracy. To overcome the challenges of cross-modal feature matching between LiDAR and camera, we introduce dual-path correspondence matching (DPCM), which leverages both structural and textural consistency for reliable 3D-2D correspondences. 
Furthermore, we formulate the calibration process as a joint optimization problem that integrates global constraints across multiple views and scenes, thereby enhancing overall accuracy.
Extensive experiments on real-world datasets demonstrate that EdO-LCEC outperforms state-of-the-art methods, particularly in scenarios involving sparse point clouds or partially overlapping sensor views.
\end{abstract}

\def\abstractname{Note to Practitioners}
\begin{abstract}
This article presents an environment-driven approach for LiDAR-camera extrinsic calibration. Unlike conventional target-free methods, the proposed EdO-LCEC not only extracts matchable features from real-world scenes but also adapts its calibration strategy based on the environmental feature density. This environmental awareness significantly enhances calibration robustness. By focusing on cross-modal feature matching and extrinsic optimization, our method performs reliably across various sensor configurations, including solid-state and mechanical LiDARs with differing fields of view and point densities. The proposed approach offers a practical and generalizable solution that improves upon existing target-free methods, facilitating deployment in sensor fusion and mechatronic systems. Our calibration software developed on EdO-LCEC will be publicly available at \url{https://mias.group/EdO-LCEC}. 
\end{abstract}


\section{Introduction}
\label{sec.intro}
\subsection{Background}
\IEEEPARstart{P}{erception} is a fundamental capability in autonomous mobile robotics, but achieving robust and generalizable perception remains a significant challenge \cite{jiao2023lce}. By allowing robots to acquire and interpret information about their surroundings, it provides the critical foundation for dependable navigation, planning, and high-level decision-making \cite{chen2024joint}. Modern LiDAR-camera fusion systems significantly enhance robotic perception capabilities \cite{pelau2021makes}. While LiDARs provide accurate spatial information, cameras capture rich textural details \cite{zhiwei2024lcec,junior2022ekf}. Their complementary fusion enables robust performance in key robotic tasks, including SLAM \cite{lin2022r,yu2022accurate, hui2025pl,tian2022kimera,shi2024fast}, object recognition \cite{qi2018frustum,wang2025segnet4d}, and localization \cite{luo2025bevplace++}. LiDAR-camera extrinsic calibration (LCEC), which estimates the relative pose between the two sensors, becomes a core and foundational process for effective data fusion in automation science and engineering. Extensive research on offline, target-based LCEC has yielded numerous effective and robust algorithms over the decades. However, online target-free methods still struggle in complex, unstructured scenes due to limited adaptability to the working environment, and much work remains to fully realize their potential.

\begin{figure*}[t!]
    \centering
    \includegraphics[width=1.0\textwidth]{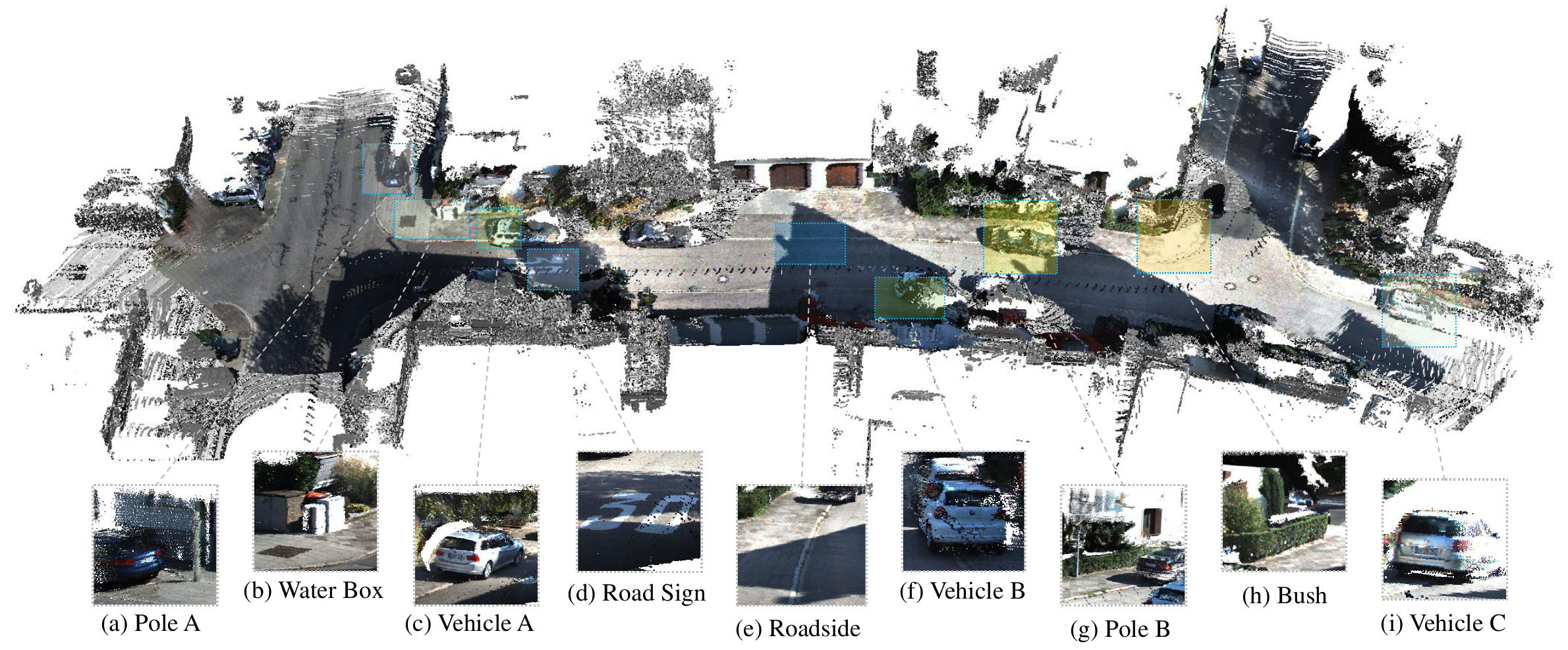}
    \caption{
    Visualization of calibration results through LiDAR and camera data fusion in KITTI odometry 00 sequence: (a)-(i) zoomed-in regions that illustrate the alignment between the camera images and the LiDAR point clouds. 
    }
    \label{fig.cover}
\end{figure*}

\subsection{Existing Challenges and Motivation}
Current LCEC methods are primarily categorized as either target-based or target-free.
Target-based approaches \cite{cui2020acsc,yan2023joint, beltran2022automatic, xie2022a4lidartag, huang2024gtscalib} have long been the preferred choice in this field. They are typically offline, relying on customized calibration targets (typically checkerboards). However, they often demonstrate poor adaptability to real-world environments. This is largely because extrinsic parameters may change significantly due to moderate shocks or during extended operations in environments with vibrations. 

Online, target-free approaches aim at overcoming this problem by extracting informative visual features directly from the environment. Previous works \cite{yuan2021pixel, pandey2015automatic} estimate the extrinsic parameters by matching the cross-modal edges between LiDAR projections and camera images.  While effective in specific scenarios with abundant features, these traditional methods heavily rely on well-distributed edge features. Recent advances in deep learning techniques have spurred extensive explorations, such as \cite{wang2022automatic, han2021auto, liao2023se}, to leverage semantic information to aid cross-modal feature matching. Although these approaches have shown compelling performance in specific scenarios, such as urban freeway, they predominantly rely on curated, pre-defined objects, \egi, vehicles, lanes, and traffic poles. 
On the other hand, several end-to-end deep learning networks \cite{borer2024chaos, lv2021lccnet,shi2020calibrcnn,zhao2021calibdnn,iyer2018calibnet,yuan2020rggnet} have been developed to find a more direct solution for LCEC. While these methods have demonstrated effectiveness on their training datasets, like KITTI \cite{geiger2012we}, they highly rely on the training setup and are thus less generalizable. {The recent work MIAS-LCEC \cite{zhiwei2024lcec} exploits the large vision model MobileSAM \cite{zhang2023faster} to improve cross-modal feature matching. While this method achieves high accuracy with solid-state LiDARs (producing dense point clouds) when the sensors have overlapping fields of view, its performance significantly degrades in challenging scenarios with sparse or incomplete point clouds.} 

Although current target-free LCEC methods eliminate reliance on calibration targets, they are still restricted to specific sensor types and pre-defined environmental settings. As a result, they are hard to balance calibration accuracy and generalizability in complex, unstructured scenes due to limited adaptability. This challenge becomes even more pronounced in dynamic scenarios where sensor poses are uncertain and frequently changing. Therefore, this study aims to create a more flexible online calibration framework that adapts its strategy based on the environmental cues, thereby improving the robustness of online, target-free LCEC.

\subsection{Novel Contributions}

In this article, we move one step forward to introduce environmental observation into target-free calibration, proposing the first environment-driven framework, EdO-LCEC. EdO-LCEC adapts to external conditions to maintain optimal performance by balancing the feature density between LiDAR projection and camera image. Unlike the conventional target-free approaches, EdO-LCEC is no longer constrained by fixed strategies but can intelligently respond to environmental changes. Specifically, as illustrated in Fig. \ref{fig.cover}, we consider the working environment of the sensors as a sequence composed of multiple scenes. 
By actively perceiving the feature density of the environment and merging multiple scenes across different times, this approach could achieve high-precision calibration dynamically. A prerequisite for the success of EdO-LCEC is the design of a generalizable scene discriminator. 
The scene discriminator employs large vision models to conduct depth estimation and image segmentation. 
In detail, it calculates the feature density of the calibration scene and uses it to guide the generation of multiple virtual cameras for projecting LiDAR intensities and depth. This improved LiDAR point cloud projecting strategy increases the available environmental features, and thus overcomes the previous reliance of algorithms \cite{yuan2021pixel,koide2023general} on uniformly distributed geometric or textural features. 
At each scene, we perform dual-path correspondence matching (DPCM). 
{Different from the C3M in MIAS-LCEC \cite{zhiwei2024lcec}, DPCM divides correspondence matching into spatial and textural pathways to fully leverage both geometric and semantic information. Each pathway constructs a cost matrix based on structural and textural consistency, guided by accurate semantic priors, to yield reliable 3D-2D correspondences.} 
Finally, the correspondences obtained from multiple views and scenes are used as inputs for our proposed {multi-view and multi-scene joint optimization}, which derives and refines the extrinsic matrix between LiDAR and camera.
Through extensive experiments conducted on three real-world datasets, EdO-LCEC demonstrates superior robustness and accuracy compared to other SoTA approaches.

To summarize, our novel contributions are as follows:
\begin{itemize}
    \item {
    EdO-LCEC, the first environment-driven, online LCEC framework that introduces environmental observation into target-free calibration.
    }
    \item {Generalizable scene discriminator, which can automatically observe the calibration scene by evaluating the feature density through potential spatial, textural, and semantic features extracted by SoTA LVMs.}
    \item {DPCM, a novel cross-modal feature matching algorithm consisting of textural and spatial pathways, capable of generating dense and reliable 3D-2D correspondences between LiDAR point cloud and camera image.}
    \item {{Multi-view and multi-scene joint relative pose optimization}, enabling high-quality extrinsic estimation {by integrating multiple perspective views within a single scene and merging distinct scenes across different timestamps}.}
\end{itemize}

\subsection{Article Structure}
The remainder of this article is structured as follows:
Sect. \ref{sec.relate_works} reviews SoTA approaches in LCEC.
Sect. \ref{sec.method} introduces EdO-LCEC, our proposed online, target-free LCEC algorithm. Sect. \ref{sec.experiment} presents experimental results and compares our method with SoTA methods.
Finally, in Sect. \ref{sec.conclusion}, we conclude this article and discuss potential future research directions.

\section{Related Work}
\label{sec.relate_works}

Target-based LCEC methods achieve high accuracy using customized calibration targets (typically checkerboards). However, they require offline execution and are significantly limited in dynamic or unstructured environments where such targets are unavailable \cite{ou2023targetless,liao2023se}. Recent studies have shifted to online, target-free approaches to overcome these limitations. Pioneering works \cite{lv2015automatic, castorena2016autocalibration, yuan2021pixel, pandey2015automatic, tang2023robust} estimate the relative pose between the two sensors by aligning the cross-modal edges or mutual information (MI) extracted from LiDAR projections and camera RGB images. While effective in specific scenarios with abundant features, these traditional methods heavily rely on well-distributed edges and rich texture, which largely compromise calibration robustness. 
To circumvent the challenges associated with cross-modal feature matching, several studies \cite{zhang2023overlap, yin2023automatic, ou2023cross, ou2023targetless} have explored motion-based methods. These approaches match sensor motion trajectories from visual and LiDAR odometry to derive extrinsic parameters through optimization. While they effectively accommodate heterogeneous sensors without requiring overlap, they demand precise synchronized LiDAR point clouds and camera images to accurately estimate per-sensor motion, which limits their applicability in real-world scenarios.
\begin{figure*}[t!]
    \centering
    \includegraphics[width=1.0\linewidth]{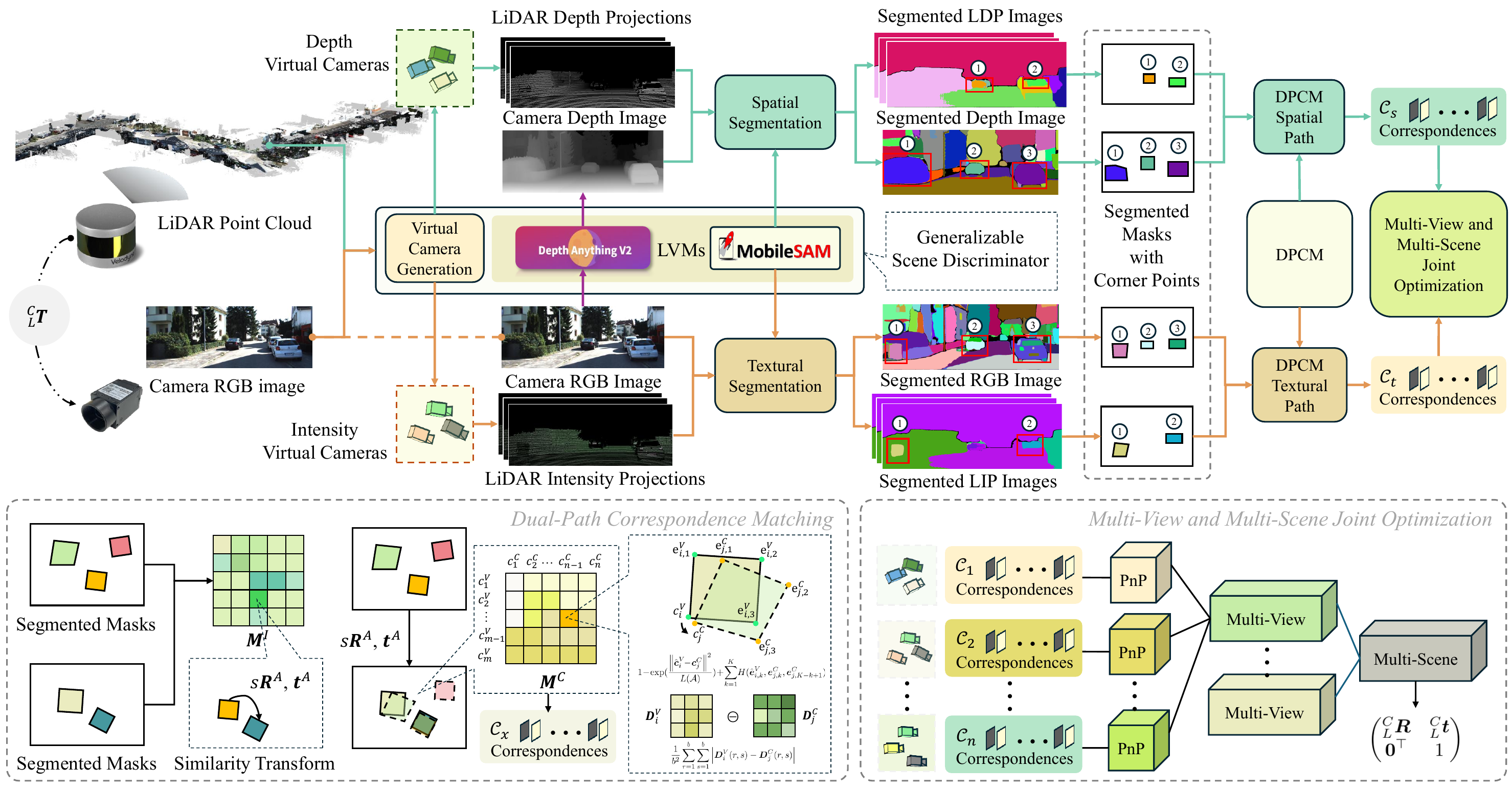}
    \caption{{The pipeline of our proposed EdO-LCEC. The working environment of the sensors is analyzed by the generalizable scene discriminator. In each calibration scene, the feature density of LiDAR and camera data is estimated by image segmentation and depth estimation. Based on this feature density, the scene discriminator generates multiple depth and intensity virtual cameras to create LIP and LDP images. Image segmentation results (segmented masks with corner points) of virtual images and camera images are sent to DPCM to obtain dense 3D-2D correspondences, which serve as input for the multi-view and multi-scene joint optimization to derive and refine the extrinsic matrix between LiDAR and camera.}
    }
    \label{fig.framework_overview}
\end{figure*}

Advances in deep learning techniques have driven significant exploration into enhancing traditional target-free algorithms. Some studies \cite{ma2021crlf, wang2022automatic, han2021auto,  koide2023general, zhiwei2024lcec} explore attaching deep learning modules to their calibration framework as useful tools to enhance calibration efficiency. For instance, \cite{ma2021crlf} accomplishes LiDAR and camera registration by aligning road lanes and poles detected by semantic segmentation. Similarly, \cite{han2021auto} employs stop signs as calibration primitives and refines results over time using a Kalman filter. A recent study \cite{koide2023general} introduced Direct Visual LiDAR Calibration (DVL), a novel point-based method that utilizes SuperGlue \cite{sarlin2020superglue} to establish direct 3D-2D correspondences between LiDAR and camera data. 
On the other hand, several learning-based algorithms \cite{lv2021lccnet,shi2020calibrcnn,zhao2021calibdnn,iyer2018calibnet,yuan2020rggnet} attempt to reformulate the calibration process into more direct solutions by leveraging end-to-end deep learning networks. Although these methods have shown promising results on public datasets such as KITTI \cite{geiger2012we}, which primarily focus on urban driving scenarios, their performance has not been extensively validated on other real-world datasets that include more diverse and challenging scenes. 
Moreover, end-to-end models trained on a single LiDAR-camera pair often overfit to the intrinsic and extrinsic parameters of the training setup and fail to generalize, even when evaluated on a different camera within the same dataset. For instance, a network trained end-to-end on the left camera of KITTI but evaluated on the right camera still predicts the pose of the left camera. This indicates that such networks primarily memorize specific extrinsic parameters rather than learning a generalizable calibration strategy. In contrast, our method incorporates feature density evaluation to achieve environment-aware calibration. This environment-driven strategy enables robots to maintain high-precision calibration across diverse conditions without being constrained by specific sensor configurations.

\section{Methodology}
\label{sec.method}

Given LiDAR point clouds and camera images, our goal is to estimate their extrinsic matrix $^{C}_{L}\boldsymbol{T}$, defined as follows:
\begin{equation}
{^{C}_{L}\boldsymbol{T}} = 
\begin{bmatrix}
{^{C}_{L}\boldsymbol{R}} & {^{C}_{L}\boldsymbol{t}} \\
\boldsymbol{0}^\top & 1
\end{bmatrix}
\in{SE(3)},
\label{eq.lidar_to_camera_point}
\end{equation}
where $^{C}_{L}\boldsymbol{R} \in{SO(3)}$ represents the rotation matrix, $^{C}_{L}\boldsymbol{t}$ denotes the translation vector, and $\boldsymbol{0}$ represents a column vector of zeros. We first give an overview of the proposed method. As shown in Fig. \ref{fig.framework_overview}, it mainly contains three stages: 
\begin{itemize}
\item{We first utilize a scene discriminator to {perceive} the environment through image segmentation and depth estimation, generating virtual cameras that project LiDAR intensity and depth from multiple viewpoints. {LiDAR projections from multiple views are further segmented into masks with corner points} (Sect. \ref{sec.scene_discriminator}).}
 \item{The segmented masks with detected corner points are processed along two pathways (spatial and textural) of the dual-path correspondence matching module to establish reliable 3D-2D correspondences (Sect. \ref{sec.dualpath_matching}).}
  \item{The obtained correspondences are used as inputs for our proposed {multi-view and multi-scene joint optimization method}, which derives and refines the extrinsic matrix {${^{C}_{L}}\boldsymbol{T}$} (Sect. \ref{sec.st_optimization}).}
\end{itemize}  

\subsection{Generalizable Scene Discriminator} 
\label{sec.scene_discriminator}
Our environment-driven approach first employs a generalizable scene discriminator to observe the surroundings by generating virtual cameras to project LiDAR point cloud intensities and depth. This discriminator configures both an intensity and a depth virtual camera from the LiDAR’s perspective. This setup yields a LiDAR intensity projection (LIP) image ${{^V_I}\boldsymbol{I}} \in{\mathbb{R}^{H\times W \times 1}}$ ($H$ and $W$ represent the image height and width) and a LiDAR depth projection (LDP) image ${{^V_D}\boldsymbol{I}}$. To align with the LDP image, the input camera RGB image ${{^C_I}\boldsymbol{I}}$ is processed using Depth Anything V2 \cite{depth_anything_v2} to obtain estimated depth images ${^C_D}\boldsymbol{I}$\footnote{In this article, the symbols in the superscript denote the type of target camera ($V$ denotes virtual camera and $C$ indicates real camera), and the subscript denotes the source of the image ($D$ is depth and $I$ is intensity).}.  To take advantage of semantic information, we utilize MobileSAM \cite{kirillov2023segment} as the image segmentation backbone. The series of $n$ detected masks in an image is defined as $\{\mathcal{M}_1, \dots, \mathcal{M}_n\}$. The corner points along the contours of masks detected are represented by $\{\boldsymbol{c}_1, \dots, \boldsymbol{c}_{m_i}\}$, where $m_i$ is the total corner points number in the $i$-th mask. An instance (bounding box), utilized to precisely fit around each mask, is centrally positioned at $\boldsymbol{o}$ and has a dimension of $h\times w$ pixels. To fully exploit the textural information, we evaluate the consistency of corner points using a texture matrix $\boldsymbol{D} \in \mathbb{R}^{b \times b}$, which encodes intensity values within a local $b \times b$ neighborhood. As depicted in Fig. \ref{fig.framework_overview}, the neighboring vertices of a corner point $\boldsymbol{c_i}$ is defined as $\{\boldsymbol{e}_{i,1}, \dots, \boldsymbol{e}_{i,K}\}$. These neighboring vertices are used to calculate structural consistency in dual-path correspondence matching.

For each virtual or camera image ${\boldsymbol{I}}$, the scene discriminator computes its feature density $\rho({\boldsymbol{I}})$, providing critical cues for feature extraction and correspondence matching. The feature density $\rho({\boldsymbol{I}})$ is defined as follows:
\begin{equation}
\rho({\boldsymbol{I}}) = \underbrace {\bigg(\log{\big(\sum_{i=1}^{n}{m_i}\big)^2}\bigg)}_{\rho_t} \underbrace {\bigg(\sum_{i=1}^{n}\log{\frac{\lvert \bigcup_{j=1}^{n} \mathcal{M}_j \rvert}{ \lvert \mathcal{M}_i\rvert} }\bigg)}_{\rho_s},
\end{equation}
where $\rho_t$ denotes the textural density and $\rho_s$ represents the structural density. The occlusion challenges caused by the different perspectives of LiDAR and the camera \cite{yuan2021pixel, zhiwei2024lcec}, combined with limited feature availability in sparse point clouds, mean that a single pair of virtual cameras is insufficient for a comprehensive view of the calibration scene. Let $E$ represent the event of capturing enough effective features, with probability $P(E) = \lambda$. If we have $n_I$ intensity virtual cameras and $n_D$ depth virtual cameras, the probability of capturing enough effective features is $1 - (1 - \lambda)^{n_I + n_D}$. In theory, as $ {n_I + n_D} \to \infty $, the probability $P(E')^{n_I + n_D} \to 0 $, leading to $1 - (1 - \lambda)^{n_I + n_D} \to 1$. Increasing the number of virtual cameras raises the likelihood of capturing more potential correspondences, thus enhancing calibration accuracy. 

However, employing an infinite number of virtual cameras during calibration is impractical. In real applications, a straightforward approach is to let users specify the number of virtual cameras based on their experience. If an initial estimate of the calibration parameters is available, the number of virtual cameras can also be inferred from the distance between the LiDAR origin and the initial estimate. Another automated option is to compute the ratio between the LiDAR and camera fields of view (FoVs) to determine the minimum number of virtual cameras required. Each virtual camera replicates the intrinsic FoV of the real camera, and together they fully cover the LiDAR scanning range. While the above strategies enhanced calibration performance in certain specific cases, they lacked generalizability across more complex and diverse scenarios. The required parameters vary with environmental changes, and the absence of environmental perception prevents the algorithm from automatically adapting to unseen and challenging conditions.

In this work, we introduce a fully automated strategy for generating virtual cameras guided by environmental information. Considering the trade-off between calibration accuracy and computational cost, we determine the number of virtual cameras to balance the feature density:
\begin{equation}
\rho({^C_I}\!\boldsymbol{I}) + \rho({^C_D}\boldsymbol{I}) = \sum_{i=0}^{n_I-1}\rho({^V_I}\!\boldsymbol{I}_{i}) + \sum_{i=0}^{n_D-1}{\rho}({^V_D}\boldsymbol{I}_i).
\end{equation}

\begin{figure}[t!]
    \centering
    \includegraphics[width=1.0\linewidth]{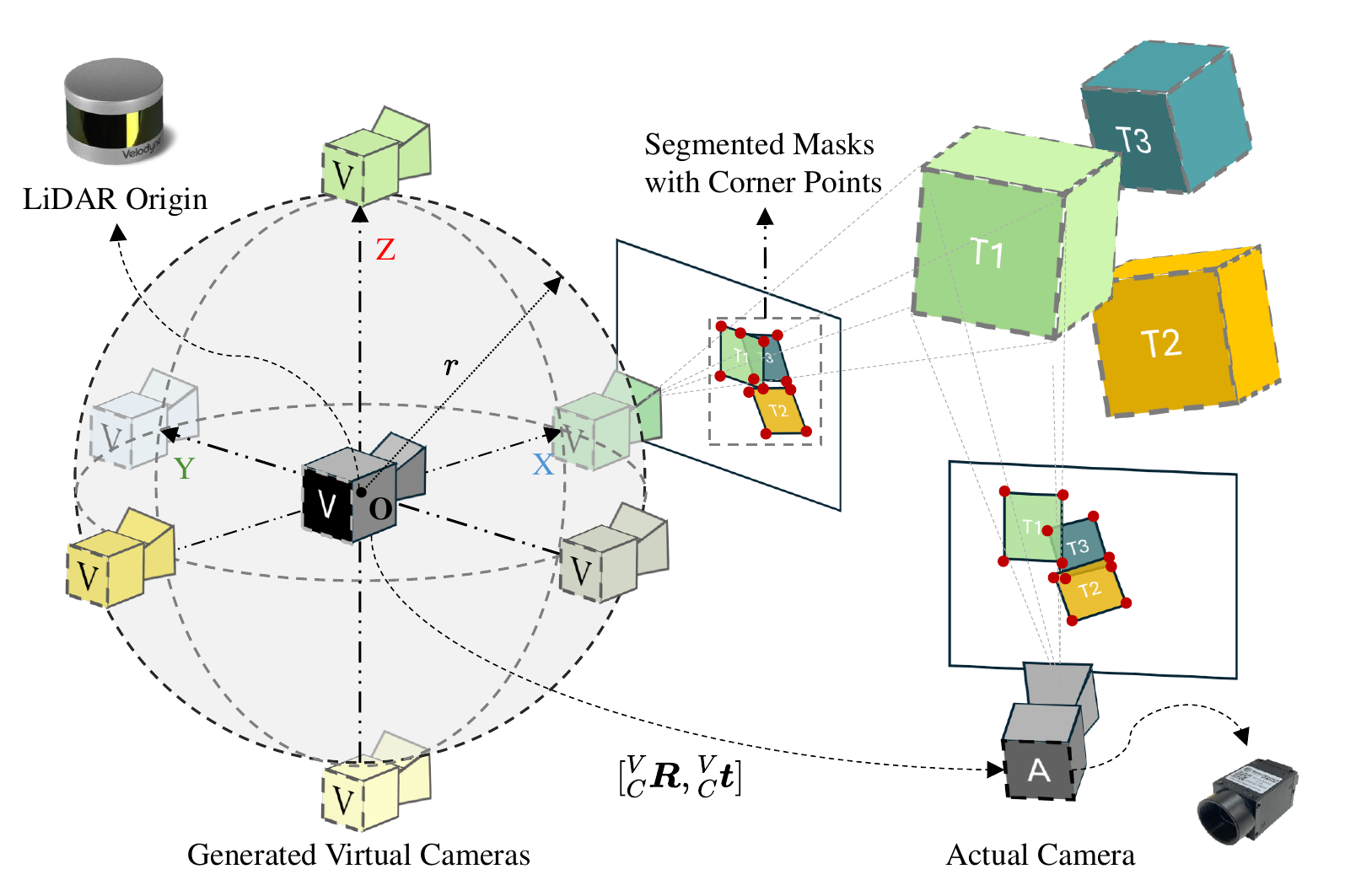}
    \caption{Virtual camera generation method. We distribute the virtual cameras along the $X$, $Y$, and $Z$ axes inside a sphere of radius $r = 0.3\,\mathrm{m}$. The scene discriminator dynamically determines the number of virtual cameras based on feature density. All cameras maintain a default front-facing orientation.
    }
    \label{fig.virutal_camera_generation}
\end{figure}

As depicted in Fig. \ref{fig.virutal_camera_generation}, in practical applications, we set multiple virtual cameras inside a sphere originating from the LiDAR perspective center. A smaller or larger sphere might be used to set more cameras if necessary.
Since the feature density is similar if the perspective viewpoints of virtual cameras are close to the initial position, we can assume that $\rho({^V_I}\!\boldsymbol{I}_{i}) \approx \rho({^V_I}\!\boldsymbol{I}_{0})$ and $\rho({^V_D}\boldsymbol{I}_{i}) \approx \rho({^V_D}\boldsymbol{I}_{0})$. So $n_I$ and $n_D$ can be obtained as follows:
\begin{equation}
n_I = \frac{\rho({^C_I}\!\boldsymbol{I})}{\rho({^V_I}\!\boldsymbol{I}_{0})}, \text{ } n_D = \frac{\rho({^C_D}\boldsymbol{I})}{\rho({^V_D}\boldsymbol{I}_{0})}.
\end{equation}
Once all virtual cameras are generated, the discriminator performs image segmentation on each LiDAR projection captured from multiple views, detecting the corner points of the masks. These masks with detected corner points serve as inputs for the dual-path correspondence matching.

\subsection{Dual-Path Correspondence Matching}
\label{sec.dualpath_matching}
\begin{figure}[t!]
    \centering
    \includegraphics[width=1.0\linewidth]{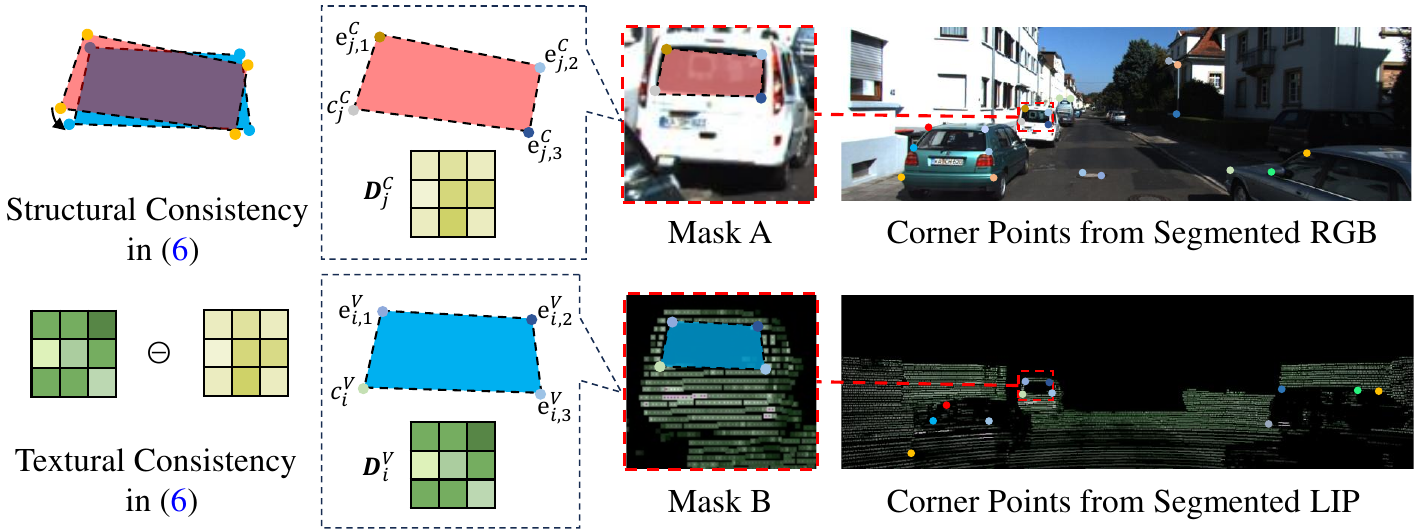}
    \caption{{DPCM utilizes structural consistency and textural consistency in (\ref{eq.adaptive_cost_func}) to compute matching cost between corner points detected from different segmented masks.}}
    \label{fig.DPCM_TS}
\end{figure}

Given the segmented masks with detected corner points, dual-path correspondence matching leverages them to achieve dense and reliable 3D-2D correspondences. {DPCM consists of two pathways, one for correspondence matching of LIP and RGB images, and the other for LDP and depth images.}  For each pathway, DPCM adopted the approach outlined in \cite{zhiwei2024lcec} to obtain mask matching result $\mathcal{A} = \{(\mathcal{M}^V_i, \mathcal{M}^C_i) \mid i = 1, \dots, m\}$ from a cost matrix $\boldsymbol{M}^I$. Each matched mask pair can estimate a 4-DoF similarity transform $[s\boldsymbol{R}^A,\boldsymbol{t}^A]$ to guide the correspondence matching. 
Specifically, we update the corner points $\boldsymbol{c}^V_i$ in the virtual image to a location $\hat{\boldsymbol{c}}^V_i$ that is close to its true projection coordinate using this affine transformation, as follows:
\begin{equation}
 \hat{\boldsymbol{c}}^V_i = s\boldsymbol{R}^A{(\boldsymbol{c}^V_i)} + \boldsymbol{t}^A.
\end{equation}
To determine optimum corner point matches, we construct a cost matrix $\boldsymbol{M}^C$, where the element at $\boldsymbol{x} = [i,j]^\top$, namely:
\begin{equation}
\begin{aligned}
&\!\boldsymbol{M}^C(\boldsymbol{x}) = \\
&\!\beta_s\!\underbrace{\bigg(1\!-  \!\exp(\frac{\left\|\hat{\boldsymbol{c}}^V_i \!\!- \!\boldsymbol{c}^C_j\right\|^2}{L(\mathcal{A})}) 
   \!+ \!{\sum_{k = 1}^{K}\!{H(\hat{\boldsymbol{e}}^V_{i,k},\boldsymbol{e}^C_{j,k},\boldsymbol{e}^C_{j,K-k+1})}}}_{\text{Structural Consistency}}\!\bigg) \\
&\!+\beta_t\underbrace{\bigg(\frac{1}{b^2}\sum_{r = 1}^{b}\sum_{s = 1}^{b}{\left |\boldsymbol{D}^V_{i}(r,s)-\boldsymbol{D}^C_{j}(r,s) \right |}\bigg)}_{\text{Textural Consistency}}
\end{aligned}
\label{eq.adaptive_cost_func}
\end{equation}
denotes the matching cost between the $i$-th corner point of a mask in the LiDAR virtual image and the $j$-th corner point of a mask in the camera image. (\ref{eq.adaptive_cost_func}) consists of structural and textural consistency. {As illustrated in Fig. \ref{fig.DPCM_TS}}, the structural consistency measures the structural difference of corner points in the virtual and real image, where $L(\mathcal{A})$ serves as a width parameter based on the average perimeter of the matched masks and $H(\hat{\boldsymbol{e}}^V_{i,k},\boldsymbol{e}^C_{j,k},\boldsymbol{e}^C_{j,K-k+1})$ represents the similarity of the neighboring vertices between current and target corner point. The textural consistency derives from the relative textural similarity of the $b$ neighboring zone. After establishing the cost matrix, a strict criterion is applied to achieve reliable matching. Matches with the lowest costs in both horizontal and vertical directions of $\boldsymbol{M}^C(\boldsymbol{x})$ are determined as the optimum corner point matches. Since every $\boldsymbol{c}^V_i$ can trace back to a LiDAR 3D point $\boldsymbol{p}^L_i = [x^L,y^L,z^L]^\top$, and every $\boldsymbol{c}^C_i$ is related to a pixel $\boldsymbol{p}_i = [u,v]^\top$ (represented in homogeneous coordinates as $\tilde{\boldsymbol{p}}_i$) in the camera image, the final correspondence matching result of DPCM is $\mathcal{C} = \{(\boldsymbol{p}^L_i,\boldsymbol{p}_i) \mid i = 1,\dots, q\}$.

\begin{algorithm}[t!]
\caption{{Dual-Path Correspondence Matching}}
\textbf{Require:}\\
Textural pathway: Segmented Masks (including detected corner points) obtained from the LIP and RGB image. \\
Spatial pathway: Segmented Masks (including detected corner points) obtained from the LDP and depth images.\\
\textbf{Stage 1 (Reliable mask matching):} \\
(1) Conduct cross-modal mask matching by adopting the method described in \cite{zhiwei2024lcec}. \\
(2) Estimate $s\boldsymbol{R}^A$ and $\boldsymbol{t}^A$ using the approach in \cite{zhiwei2024lcec}.\\
\textbf{Stage 2 (Dense correspondence matching):} \\
(1) For each pathway, update all masks in the virtual image using $s\boldsymbol{R}^A$ and $\boldsymbol{t}^A$.\\
(2) Construct the corner point cost matrix $\boldsymbol{M}^C$ using (\ref{eq.adaptive_cost_func}). \\
(3) Select matches with the lowest costs in both horizontal and vertical directions of $\boldsymbol{M}^{C}(\boldsymbol{x})$ as the optimum correspondence matches. \\
(4) Aggregate all corner point correspondences to form the sets $\mathcal{C} = \{(\boldsymbol{p}^L_i,\boldsymbol{p}_i) \mid i = 1,\dots, q\}$.
\label{Algorithm.DPCM}
\end{algorithm}
{Leveraging the generalizable scene discriminator, the multi-perspective views generated by virtual cameras provide rich spatial-textural descriptors that effectively transform the feature matching problem into a robust 3D-2D correspondence establishment process. DPCM achieves efficient computation by verifying 2D structural and textural consistency instead of performing costly 3D point cloud registration.
The pseudo-code of our DPCM is presented in Algorithm \ref{Algorithm.DPCM}. During corner point matching within a mask, while the C3M in MIAS-LCEC exclusively considers points from the corresponding mask (\iei, the matched mask associated with the current corner point), our DPCM in EdO-LCEC incorporates additional corner points from neighboring masks. This innovative matching strategy offers two key advantages: (1) it partially eliminates the dependency on dense, high-precision mask matching results, and (2) it substantially avoids errors induced by imperfect mask matching, especially under challenging conditions such as sparse LiDAR point clouds or limited LiDAR-camera FoV overlap.}

\subsection{Multi-View and Multi-Scene Joint Optimization}
\label{sec.st_optimization}

{EdO-LCEC models the sensor’s operational environment as a composition of $N$ distinct scenes across different timestamps. While single-view methods (\egi, \cite{yuan2021pixel,zhiwei2024lcec}) suffer from limited high-quality correspondences in sparse or incomplete point cloud scenarios, our environment-driven approach overcomes this constraint through integrating observations from multiple views and scenes. After establishing 3D-2D correspondences by DPCM, EdO-LCEC computes the extrinsic matrix between LiDAR and camera through a multi-view and multi-scene joint optimization. By aggregating optimal correspondences across time and space, our method enhances matching robustness, maximizes high-quality feature associations, and ultimately improves calibration accuracy.}

{In multi-view optimization}, the extrinsic matrix ${^{C}_{L}\hat{\boldsymbol{T}}_t}$ of the $t$-th scene can be formulated as follows:
\begin{equation}
{^{C}_{L}\hat{\boldsymbol{T}}_t}\! =\! \underset{^C_L{\boldsymbol{T}_{t,k}}}{\arg\min} \!\!\!\!
\sum_{i = 1}^{n_I + n_D}\!\!\!\sum_{(\boldsymbol{p}^L_j,\boldsymbol{p}_j)\in\mathcal{C}_i}\!\!\!\!\!{G\bigg(\underbrace {\left\|\pi(
{^{C}_{L}\boldsymbol{T}_{t,k}}
 \boldsymbol{p}_{j}^L) - {\tilde{\boldsymbol{p}}}_{j}\right\|_2}_{\epsilon_j}\bigg)}, 
\label{eq.multi_view}
\end{equation}
where ${^{C}_{L}\boldsymbol{T}_{t,k}}$ denotes the $k$-th PnP solution obtained using a selected subset $\mathcal{V}_{k}$ of correspondences from $\mathcal{C}_i$, and $\epsilon_j$ represents the reprojection error of $(\boldsymbol{p}^L_i,\boldsymbol{p}_i)$ with respect to ${^{C}_{L}\boldsymbol{T}_{t,k}}$. $G(\epsilon_j)$ represents the gaussian depth-normalized reprojection error under the projection model $\pi$, defined as:
\begin{equation}
G(\epsilon_j) = \frac{\epsilon_j(e-\mathcal{K}(d'_{j},\bar{d'}))}
{H + \epsilon_j},
\label{eq.dpcm_gaussian}
\end{equation}
where ${d'_j}$ is the normalized depth of $\boldsymbol{p}_{j}^L$, $\bar{d'}$ is the average normalized depth, and $\mathcal{K}(d'_{j},\bar{d'})$ is a gaussian kernal. 

In multi-scene optimization, we choose a reliable subset $\mathcal{S}_t$ from each scenario, which can be obtained as follows:
\begin{equation}
    \mathcal{S}_t = \bigcup_{k=1}^{s_t}{\mathcal{V}_{t,k}}, \text{ } s_t = \min\{\frac{Q_{\max}q_t}{\sum^{N}_{j = 1}{q_j}}, s_{\max}\},
\end{equation}
where ${\mathcal{V}_{t,k}}$ is the $k$-th selected correspondences subset in the $t$-th scene. $q_t$ denotes the number of correspondences in the $t$-th scene. $Q_{\max}$ represents the maximum number of correspondences hope to get from all scenarios, and $s_{\max}$ denotes the maximum number of reliable correspondences in a single scene. The multi-scene optimization process then solves the final extrinsic matrix ${^{C}_{L}{\boldsymbol{T}}^*}$ by minimizing the joint reprojection error:
\begin{equation}
\mathcal{L}({^{C}_{L}{\boldsymbol{T}}^{*}}) = 
\sum_{t = 1}^{N} {\sum_{(\boldsymbol{p}^L_{j},\boldsymbol{p}_{j})\in\mathcal{S}_t}\!{G\bigg(\left\|\pi(
{^{C}_{L}\boldsymbol{T}^{*}}
 \boldsymbol{p}_{j}^L) - {\tilde{\boldsymbol{p}}}_{j}\right\|_2}\bigg)}
 \label{eq.multi_scene}
\end{equation}
across multiple spaces in the environment at different times. This process combines optimal correspondences from both spatial and textural pathways in each scenario, enabling robust environment-driven optimization through multi-view and multi-scene fusion. As a result, our method achieves human-like adaptability to dynamic environments. {This environment perception makes EdO-LCEC behave much better than other target-free approaches, especially in conditions when point clouds are sparse or incomplete}.

\section{Experiment}
\label{sec.experiment}

\subsection{Experimental Setup and Evaluation Metrics}
\label{sec.exp_setup}

\begin{table*}[t!]
\caption{Quantitative comparisons with SoTA target-free LCEC approaches on the 00 sequence of KITTI odometry. The best results are shown in bold type. \dag: These methods did not release code, preventing the reproduction of results for both cameras.}
\centering
\fontsize{6.5}{10}\selectfont
\begin{tabular}{l|c|c@{\hspace{0.15cm}}c|c@{\hspace{0.15cm}}c@{\hspace{0.15cm}}c|c@{\hspace{0.15cm}}c@{\hspace{0.15cm}}c|c@{\hspace{0.15cm}}c|c@{\hspace{0.15cm}}c@{\hspace{0.15cm}}c|c@{\hspace{0.15cm}}c@{\hspace{0.15cm}}c}
\toprule
\multirow{3}*{Approach}& \multirow{3}*{Initial Range}&\multicolumn{8}{c|}{Left Camera} &\multicolumn{8}{c}{Right Camera}\\
\cline{3-18}
&&\multicolumn{2}{c|}{Magnitude}
&\multicolumn{3}{c|}{Rotation Error ($^\circ$)} &\multicolumn{3}{c|}{Translation Error (m)} 
&\multicolumn{2}{c|}{Magnitude}
&\multicolumn{3}{c|}{Rotation Error ($^\circ$)} &\multicolumn{3}{c}{Translation Error (m)}\\

&& $e_r$ ($^\circ$) & $e_t$ (m) & Yaw & Pitch & Roll  & {X} &  {Y} &  {Z}   & $e_r$ ($^\circ$) & {$e_t$ (m)} & Yaw & Pitch & Roll   &  {X} &  {Y} &  {Z}\\
\hline
\hline
CalibRCNN\textsuperscript{\dag} \cite{shi2020calibrcnn} &$\pm10^\circ / \pm 0.25m$ &0.805 &0.093	&0.446	&0.640	&0.199	&0.062	&0.043 &0.054 &- &-	&-	&-	&-	&-	&-	&-\\
CalibDNN\textsuperscript{\dag} \cite{zhao2021calibdnn} &$\pm10^\circ / \pm 0.25m$ &1.021 &0.115	&0.200	&0.990	&0.150	&0.055	&0.032	&0.096 &- &-&-	&-	&-	&-	&-	&-\\
RegNet\textsuperscript{\dag} \cite{schneider2017regnet} &$\pm20^\circ / \pm 1.5m$ &0.500 &0.108	&0.240	&0.250	&0.360	&0.070	&0.070	&0.040 &- &-&-	&-	&-	&-	&-	&-\\
LCCNet \cite{lv2021lccnet} &$\pm10^\circ / \pm 1.0m$ &1.418
&0.600  &0.455 &0.835 &0.768 &0.237 &0.333  &0.329 &1.556  &0.718 & 0.457 &1.023 &0.763 &0.416 &0.333 &0.337 \\
RGGNet \cite{yuan2020rggnet} &$\pm20^\circ / \pm 0.3m$ &1.290 &0.114	&0.640 &0.740&0.350 &0.081 &\textbf{0.028} &0.040 &3.870 &0.235 &1.480 &3.380 &0.510	&0.180 &\textbf{0.056} &0.061 \\
CalibNet \cite{iyer2018calibnet} &$\pm10^\circ / \pm 0.2m$ &5.842 &0.140	&2.873 &2.874 &3.185 &0.065 &0.064 &0.083 &5.771 &0.137	&2.877 &2.823 &3.144 &\textbf{0.063} &0.062 &0.082\\
\hline
Borer \etal\textsuperscript{\dag} \cite{borer2024chaos}&$\pm1^\circ / \pm 0.25m$  &0.455 &0.095	&0.100 &0.440 &\textbf{0.060}  &\textbf{0.037} &0.030 &0.082 &- &- &-	&-	&-	&-	&-	&- \\
CRLF \cite{ma2021crlf} &- &0.629	&4.116	&\textbf{0.033}	&0.464	&0.416	&3.648	&1.473	&0.550 &0.633	&4.606	&\textbf{0.039}	&0.458	&0.424	&4.055	&1.636	&0.644\\
UMich \cite{pandey2015automatic} &- &4.161	&0.319	&0.113	&3.111	&2.138	&0.286	&0.068	&0.086
 &4.285&0.329	&0.108	&3.277	&2.088	&0.290	&0.085	&0.090\\
HKU-Mars \cite{yuan2021pixel} &- &33.84	&6.354	&19.89	&18.71	&19.32	&3.353	&3.228	&2.419
&32.89	&4.913	&18.99	&15.77	&17.00	&2.917	&2.564	&1.646\\
DVL \cite{koide2023general}  &-&{122.1} &{5.129}	&{48.64}	&{87.29}	&{98.15}	&{2.832}	&{2.920}	&{1.881} &{120.5} &{4.357}	&{49.60}	&{87.99}	&{96.72}	&{2.086}	&{2.517}	&{1.816}\\
MIAS-LCEC \cite{zhiwei2024lcec}  &-&5.385	&1.013	&1.574	&4.029	&4.338	&0.724	&0.373	&0.343 &7.655	&1.342	&1.910	&5.666	&6.154	&0.843	&0.730	&0.358\\
\hline
\textbf{EdO-LCEC (Ours)}  &- &\textbf{0.295}	&\textbf{0.078}	&0.117	&\textbf{0.176}	&{0.150}	&0.051	&0.038	&\textbf{0.032} 
&\textbf{0.336}	&\textbf{0.118}	&0.216	&\textbf{0.168}	&\textbf{0.121}	&0.083	&0.067	&\textbf{0.032}
\\
\bottomrule
\end{tabular}
\label{tab.rescmp_kitti00}
\end{table*}

\begin{table*}[t!]
\caption{Comparisons with SoTA LCEC approaches on KITTI odometry (01-09 sequences). The best results are shown in bold type. }
\centering
\fontsize{6.7}{10}\selectfont
\begin{tabular}{l|c@{\hspace{0.15cm}}c|c@{\hspace{0.15cm}}c|c@{\hspace{0.15cm}}c|c@{\hspace{0.15cm}}c|c@{\hspace{0.15cm}}c|c@{\hspace{0.15cm}}c|c@{\hspace{0.15cm}}c|c@{\hspace{0.15cm}}c|c@{\hspace{0.15cm}}c}
\toprule
\multirow{2}*{Approach}& \multicolumn{2}{c|}{01} & \multicolumn{2}{c|}{02} & \multicolumn{2}{c|}{03}  & \multicolumn{2}{c|}{04} & \multicolumn{2}{c|}{ 05} &  \multicolumn{2}{c|}{06}&  \multicolumn{2}{c|}{07} & \multicolumn{2}{c|}{08} & \multicolumn{2}{c}{09}\\
\cline{2-19}
& $e_r$ & $e_t$& $e_r$ & $e_t$ & $e_r$ & $e_t$  & $e_r$ & $e_t$ & $e_r$ & $e_t$ & $e_r$ & $e_t$ & $e_r$  & $e_t$ & $e_r$  & $e_t$ & $e_r$  & $e_t$\\
\hline
\hline
CRLF \cite{ma2021crlf}  
&\textbf{0.623}	&7.363
&0.632	&3.640
&0.845	&6.007
&\textbf{0.601}	&0.372
&0.616	&5.961
&0.615	&25.762
&0.606	&1.807
&0.625	&5.376
&0.626	&5.133\\
UMich \cite{pandey2015automatic}  &2.196	&\textbf{0.303} &3.733
&0.329	&3.201&0.316 &2.086&0.348	&3.526
&0.356 &2.914&0.353	&3.928&0.368 &3.722
&0.367 
&3.117	&0.363\\
HKU-Mars \cite{yuan2021pixel}  &20.73	&3.768
&32.95	&12.69
&21.99	&3.493
&4.943	&0.965
&34.42	&6.505
&25.20	&7.437
&33.10	&7.339
&26.62	&8.767
&20.38	&3.459\\
DVL \cite{koide2023general}  &112.0&2.514
&120.6&4.285	
&124.7&4.711
&113.5&4.871
&123.9&4.286
&128.9&5.408	
&124.7&5.279 
&126.2&4.461 
&116.7	&3.931\\

MIAS-LCEC \cite{zhiwei2024lcec}  
&0.621	&0.298
&0.801	&0.324
&1.140	&0.324
&0.816	&0.369
&4.768	&0.775
&2.685	&0.534
&11.80	&1.344
&5.220	&0.806
&0.998	&0.432
\\
\textbf{EdO-LCEC (Ours)} &2.269 &0.462
 &\textbf{0.561}&\textbf{0.137}
&\textbf{0.737}&\textbf{0.137} &1.104&\textbf{0.339}
&\textbf{0.280}&\textbf{0.093} &\textbf{0.485}&\textbf{0.124} &\textbf{0.188}&\textbf{0.076} &\textbf{0.352}&\textbf{0.115}
&\textbf{0.386}	&\textbf{0.120}\\
\bottomrule
\end{tabular}
\label{tab.rescmp_kitti_01_08}
\end{table*}

\begin{table*}[t!]
\caption{Comparisons with SoTA target-free LCEC approaches on MIAS-LCEC-TF70. The best results are shown in bold type.}
\centering
\fontsize{6.7}{10}\selectfont
\begin{tabular}{l|cc|cc|cc|cc|cc|cc|cc}
\toprule
\multirow{2}*{Approach}& \multicolumn{2}{c|}{\makecell{Residential \\ Community}} & \multicolumn{2}{c|}{Urban Freeway} & \multicolumn{2}{c|}{Building}  & \multicolumn{2}{c|}{\makecell{Challenging \\Weather}} & \multicolumn{2}{c|}{Indoor} &  \multicolumn{2}{c|}{\makecell{Challenging \\Illumination}} &  \multicolumn{2}{c}{All} \\
\cline{2-15}
& $e_r$ ($^\circ$) & $e_t$ (m)& $e_r$ ($^\circ$) & $e_t$ (m)& $e_r$ ($^\circ$) & $e_t$ (m)& $e_r$ ($^\circ$) & $e_t$ (m)& $e_r$ ($^\circ$) & $e_t$ (m)& $e_r$ ($^\circ$) & $e_t$ (m)& $e_r$ ($^\circ$) & $e_t$ (m)\\
\hline
\hline
CRLF \cite{ma2021crlf}  &1.594&0.464
&1.582&0.140 &1.499 &20.17 &1.646 &2.055
&1.886 &30.05 &1.876 &19.05 &1.683 &11.13 \\
UMich \cite{pandey2015automatic}  &4.829
&0.387 &2.267&0.166 &11.914 &0.781 &1.851 &0.310 &2.029  &0.109 &5.012 &0.330 &4.265 &0.333\\
HKU-Mars \cite{yuan2021pixel}  &2.695
&1.208 &2.399 &1.956 &1.814 &0.706 &2.578 &1.086  &2.527 &0.246 &14.996 &3.386 &3.941 &1.261
 \\
DVL \cite{koide2023general}  &0.193
&0.063
&0.298
&0.124
&0.200
&0.087
&0.181
&0.052
&0.391
&0.030
&1.747
&0.377
&0.423
&0.100 \\

MIAS-LCEC \cite{zhiwei2024lcec}  &0.190 &0.050 &\textbf{0.291} &0.111
&0.198&0.072&\textbf{0.177}&0.046&0.363&\textbf{0.024}&0.749&0.118&0.298&0.061 \\
\textbf{EdO-LCEC (Ours)} &\textbf{0.168} &\textbf{0.044} &0.293 &\textbf{0.105} &\textbf{0.184} &\textbf{0.057}  &0.183 &\textbf{0.044} &\textbf{0.338} &0.027 &\textbf{0.474} &\textbf{0.104} &\textbf{0.255} &\textbf{0.055} \\
\bottomrule
\end{tabular}
\label{tab.rescmp_mias_tf70}
\end{table*}

In our experiments, we evaluate our proposed EdO-LCEC on four public datasets: KITTI odometry \cite{geiger2012we} (including 00-09 sequences), KITTI360 \cite{liao2022kitti}, nuScenes \cite{caesar2020nuscenes}, and MIAS-LCEC \cite{zhiwei2024lcec} (including target-free datasets MIAS-LCEC-TF70 and MIAS-LCEC-TF360). Extensive comparisons with SoTA LCEC approaches and abundant ablation studies are conducted to comprehensively validate each component of the proposed algorithm.
Following previous work \cite{lv2021lccnet,zhiwei2024lcec,iyer2018calibnet,koide2023general}, we utilize the average magnitude $e_r$ of Euler angle error and the average magnitude $e_t$ of the translation error to quantify the calibration errors. Each axes of the Euler angle error and translation error are also provided to comprehensively demonstrate the calibration accuracy.

Notably, sequences in KITTI odometry, aside from 00, were included in the training datasets for the learning-based methods \cite{lv2021lccnet,yuan2020rggnet,zhao2021calibdnn}. To ensure a fair comparison, we reproduced calibration results for both the left and right cameras on sequence 00 when the authors provided their code; otherwise, we used the reported results for the left camera from their papers. Since most learning-based methods lack APIs for custom data, our comparison with these methods is limited to the KITTI odometry 00 sequence. In nuScenes, the point clouds are captured by a 32-line spinning LiDAR (Velodyne HDL32E), which makes them extremely sparse for previous calibration approaches. To the best of our knowledge, few LCEC methods have been evaluated on this dataset. Following the official split, we use the scenes in v1.0-test of nuScenes to construct the evaluation data. For experiments on the MIAS-LCEC dataset, as the results for the compared methods \cite{ma2021crlf,yuan2021pixel,koide2023general,zhiwei2024lcec,pandey2015automatic} are reported in \cite{zhiwei2024lcec}, we directly use the values presented in that paper.

Our algorithm was implemented on an Intel i7-14700K CPU and an NVIDIA RTX4070Ti Super GPU. The entire process of a single-view calibration, including scene discriminating, DPCM, and relative pose optimization, takes approximately 15 to 70 seconds.

\begin{table}[t!]
\caption{
Quantitative comparisons of our proposed EdO-LCEC approach with other SoTA target-free approaches on the MIAS-LCEC-TF360. The best results are shown in bold type.}
\centering
\settablefont
\begin{tabular}{l|rr|rr}
\toprule
\multirow{2}*{Approach} & \multicolumn{2}{c|}{Indoor} & \multicolumn{2}{c}{Outdoor}\\
 & $e_r$ ($^\circ$) & $e_t$ (m)  & $e_r$ ($^\circ$) & $e_t$ (m)\\
\hline
\hline
CRLF \cite{ma2021crlf}  &1.479	&13.241	&1.442	&0.139\\
UMich \cite{pandey2015automatic} &1.510	&0.221	&6.522	&0.269\\
HKU-Mars \cite{yuan2021pixel}  &85.834	&7.342	&35.383	&8.542\\
DVL \cite{koide2023general} &39.474	&0.933	&65.571	&1.605\\
MIAS-LCEC \cite{zhiwei2024lcec} &0.996	&0.182	&0.659	&0.114\\
\textbf{EdO-LCEC (Ours)} &\textbf{0.720}&\textbf{0.106}	&\textbf{0.349}	&\textbf{0.109}
\\
\bottomrule
\end{tabular}
\label{tab.exp_mid360}
\end{table}

\begin{table}[t!]
\caption{
Quantitative comparisons of our proposed EdO-LCEC approach with other SoTA target-free approaches on the nuScenes dataset. The best results are shown in bold type.}
\centering
\fontsize{6.5}{10}\selectfont
\begin{tabular}{l|c@{\hspace{0.15cm}}c|c@{\hspace{0.15cm}}c@{\hspace{0.15cm}}c|c@{\hspace{0.15cm}}c@{\hspace{0.15cm}}c}
\toprule
\multirow{2}*{Approach} & \multicolumn{2}{c|}{Magnitude} & \multicolumn{3}{c|}{Rotation Error ($^\circ$)} & \multicolumn{3}{c}{Translation Error (m)}\\
 &$e_r$ ($^\circ$) & $e_t$ (m) & Yaw & Pitch & Roll  & {X} &  {Y} &  {Z}\\
\hline
\hline
CRLF \cite{ma2021crlf}&2.446	&14.82	&0.537	&0.454	&2.317	&12.25	&4.159	&4.852
\\
UMich \cite{pandey2015automatic} &4.340	&0.390	&0.524	&2.118	&3.185	&0.297	&0.071	&0.200
\\
HKU-Mars \cite{yuan2021pixel}  &36.69	&17.55	&16.150	&24.91	&19.69	&11.216	&5.878	&7.171
\\
DVL \cite{koide2023general}  &108.9	&3.371	&42.50	&71.71	&70.88	&1.672	&1.874	&1.380
\\
MIAS-LCEC \cite{zhiwei2024lcec}  &2.731	&0.688	&0.625	&0.463	&2.570	&0.554	&\textbf{0.072}	&0.325
\\
\textbf{EdO-LCEC (Ours)}  &\textbf{1.090} &\textbf{0.261} &\textbf{0.414} &\textbf{0.356} &\textbf{0.842} &\textbf{0.202} &0.076 &\textbf{0.100}\\							

\bottomrule
\end{tabular}
\label{tab.exp_nuScenes}
\end{table}

\subsection{Comparison with State-of-the-Art Method}
\label{sec.exp_dataset}

In this section, quantitative comparisons with SoTA approaches on three datasets are presented in  Fig. \ref{fig.exp_on_different_density}, Tables \ref{tab.rescmp_kitti00}
\footnote{The reproduced results of LCCNet yield higher errors compared to those reported in their paper.}, \ref{tab.rescmp_kitti_01_08}, \ref{tab.rescmp_mias_tf70}, and \ref{tab.exp_mid360}. Additionally, qualitative results are illustrated in Figs. \ref{fig.datafusion_on_checkerboard} and \ref{fig.SoTA_visualization}. 

\subsubsection{Evaluation on KITTI Odometry}
\begin{figure*}[t!]
    \centering
    \includegraphics[width=1.0\linewidth]{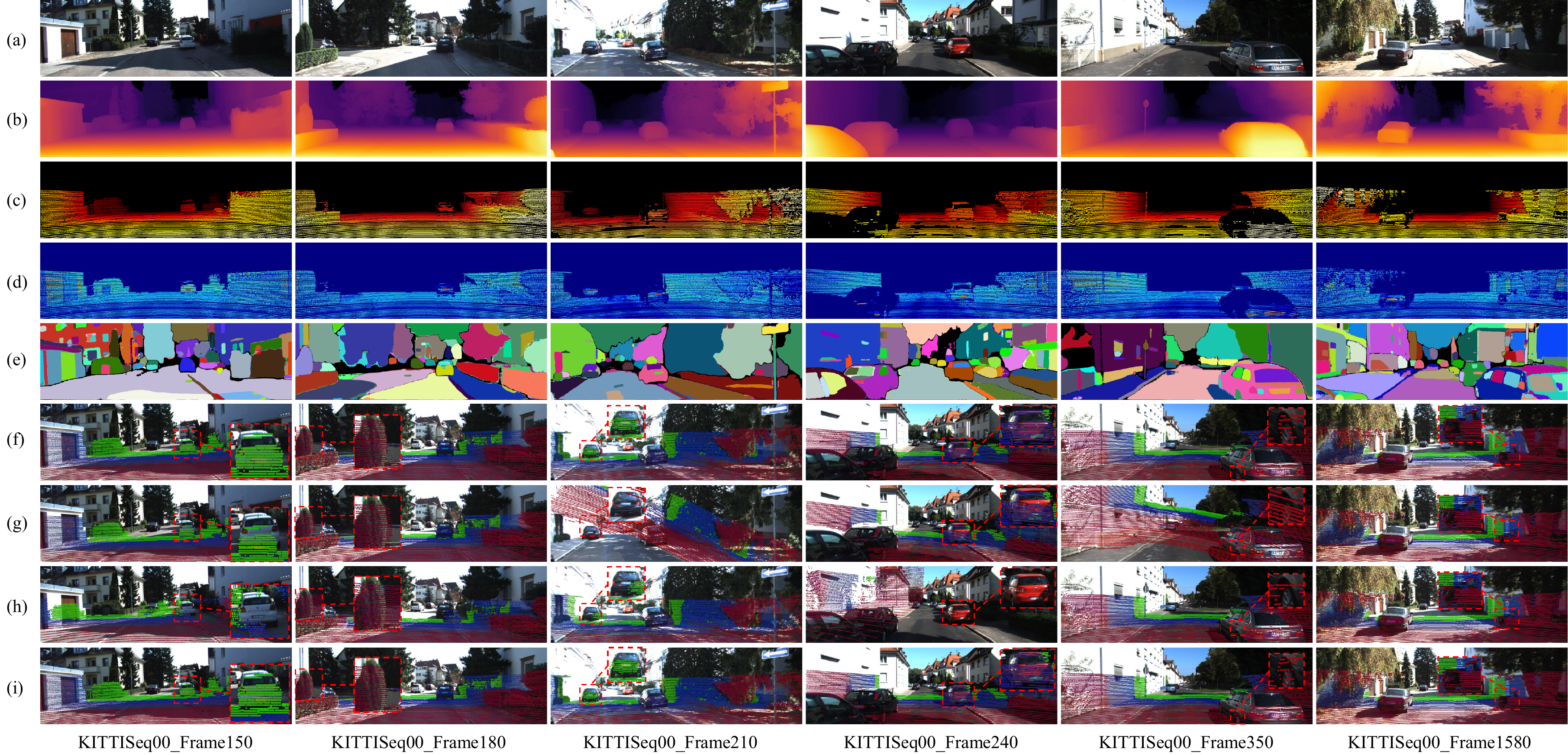}
    \caption{Qualitative comparisons with SoTA target-free LCEC approaches on the KITTI odometry dataset: (a)-(e) RGB images, Depth images, LDP images, LIP images and image segmentation results; (f)-(i) experimental results achieved using ground truth, UMich, HKU-Mars and EdO-LCEC (ours), shown by merging LiDAR depth projections and RGB images, where significantly improved regions are shown with red dashed boxes.}
    \label{fig.SoTA_visualization}
\end{figure*}

The results shown in Table \ref{tab.rescmp_kitti00} and \ref{tab.rescmp_kitti_01_08} suggest that, with the exception of sequences 01 and 04, our method achieves SoTA performance across the ten sequences (00-09) in KITTI odometry. Specifically, in the 00 sequence, EdO-LCEC reduces the $e_r$ by around 35.2-99.8\% and the $e_t$ by 16.1-98.8\% for the left camera, and reduces the $e_r$ by around 46.9-99.7\% and the $e_t$ by 13.9-97.6\% for the right camera. Additionally, according to Fig. \ref{fig.SoTA_visualization}, it can be observed that the point cloud of a single frame in KITTI is so sparse that the other approaches behave poorly. In contrast, our proposed method overcomes this difficulty and achieves high-quality data fusion through the calibration result. We attribute these performance improvements to our multi-view and multi-scene joint optimization. Merging the optimal matching results from multiple views and scenes maximizes the number of reliable correspondences and ultimately improves overall calibration accuracy.

\subsubsection{Evaluation on MIAS-LCEC Dataset}
\label{sec.exp_cmp_eva_mias_lcec}
\begin{figure}[t!]
    \centering
    \includegraphics[width=1.0\linewidth]{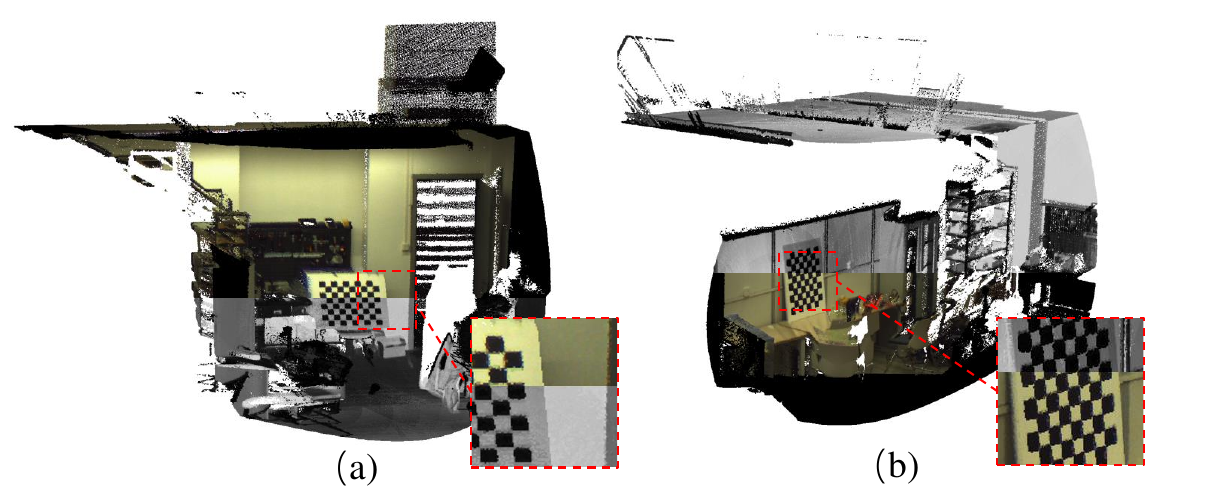}
    \caption{Visualization of EdO-LCEC calibration results through LiDAR and camera data fusion: (a)-(b) illustrate two LiDAR point clouds in MIAS-LCEC-TF70, partially rendered by the image color using the estimated extrinsic matrix of EdO-LCEC. }
    \label{fig.datafusion_on_checkerboard}
\end{figure}

Compared to the sparse point clouds in KITTI odometry, the point clouds in the MIAS-LCEC datasets are significantly denser, which facilitates feature matching and allows us to test the upper limits of the algorithm calibration accuracy. 
The results shown in Table \ref{tab.rescmp_mias_tf70} demonstrate that our method outperforms all other SoTA approaches on MIAS-LCEC-TF70. It can also be observed that our method dramatically outperforms CRLF, UMich, DVL, HKU-Mars, and is slightly better than MIAS-LCEC across the total six subsets. In challenging conditions that are under poor illumination and adverse weather, or when few geometric features are detectable, EdO-LCEC performs significantly better than all methods, particularly. This impressive performance can be attributed to the generalizable scene discriminator. The multiple virtual cameras generated by the scene discriminator provide a comprehensive perception of the calibration scene from both spatial and textural perspectives, which largely increases the possibility of capturing high-quality correspondences for the PnP solver. Furthermore, the data fusion results in Fig. \ref{fig.datafusion_on_checkerboard}, obtained using our optimized extrinsic matrix, visually demonstrate perfect alignment on the checkerboard. This highlights the high calibration accuracy achieved by our method.
\begin{figure}[t!]
    \centering
    \includegraphics[width=1.0\linewidth]{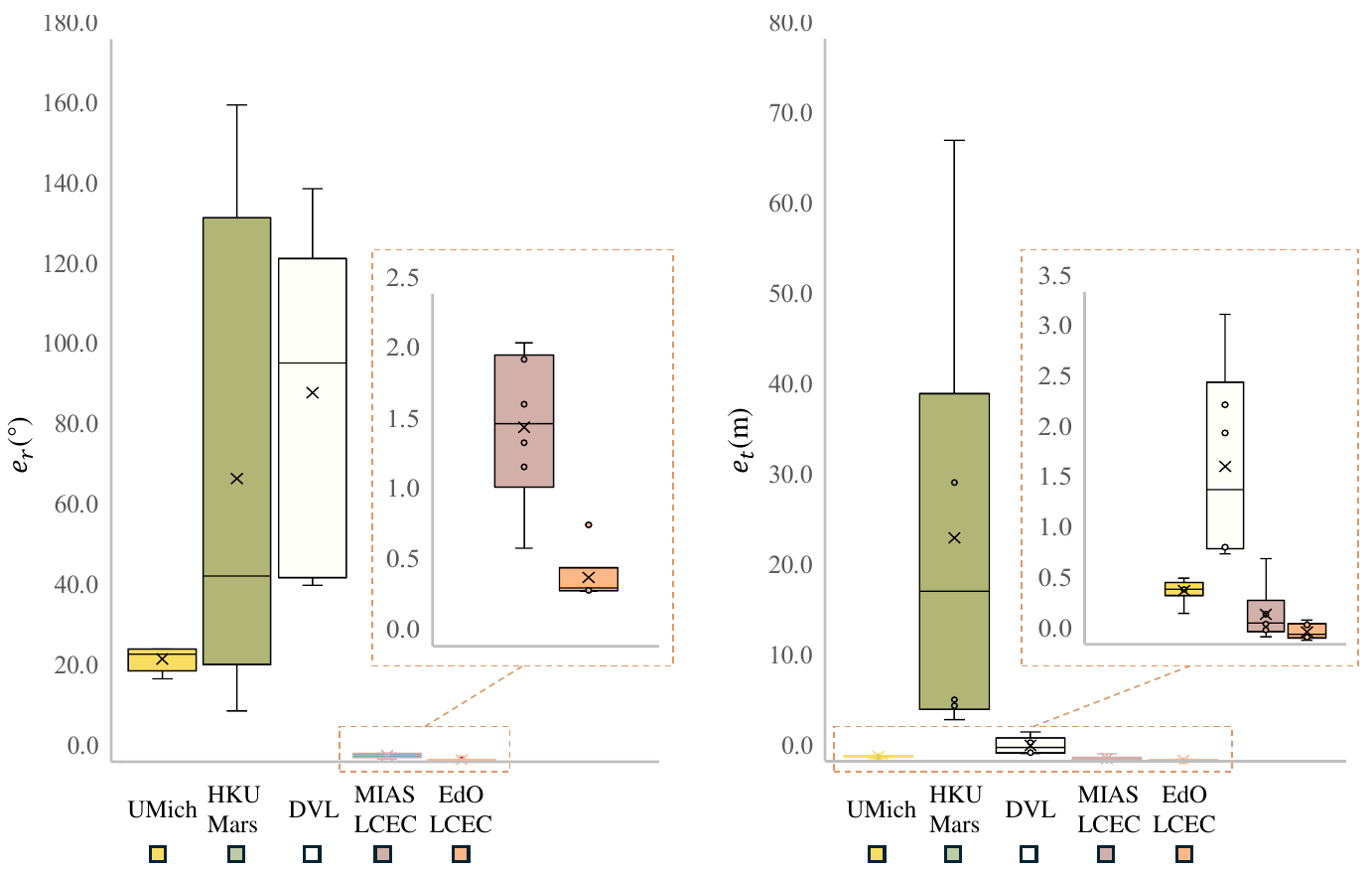}
    \caption{Comparisons with SoTA approaches on the segmented point clouds from MIAS-LCEC-TF360. The zoomed-in region highlights the comparative details between the algorithms with higher accuracy. Since the results of CRLF are invalid, they were not included in the comparison.}
    \label{fig.exp_on_different_density}
\end{figure}
\begin{figure*}[t!]
    \centering
    \includegraphics[width=1.0\linewidth]{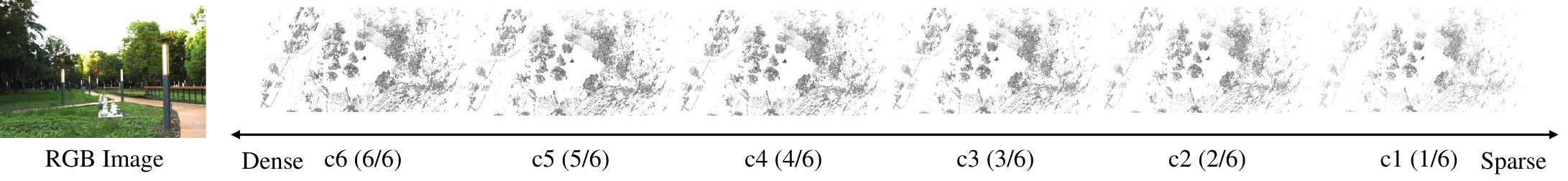}
    \caption{We divided each point cloud into six equal parts and combined these segments to create point clouds with varying densities.}
    \label{fig.visualization_sparse_pcloud}
\end{figure*}
Additionally, experimental results on the MIAS-LCEC-TF360 further prove our outstanding performance. From Table \ref{tab.exp_mid360}, it is evident that while the other approaches achieve poor performances, our method demonstrates excellent accuracy, indicating strong adaptability to more challenging scenarios, with narrow overlapping areas between LiDAR projections and camera images. This impressive performance can be attributed to our proposed DPCM, a powerful cross-modal feature matching algorithm. DPCM utilizes structural and textural consistency to jointly constrain correspondences matching on both spatial and textural pathways. This largely increases reliable correspondences compared to DVL and MIAS-LCEC, thereby providing a more reliable foundation for extrinsic parameter optimization. 

\subsubsection{Comparison with SoTA Approaches on nuScenes}
\label{sec.cmp_on_nuscenes}
Following the official split of nuScenes, we adopt the 150 test scenes in the v1.0-test set to construct the evaluation data. The experimental results, summarized in Table \ref{tab.exp_nuScenes}, show that EdO-LCEC significantly outperforms prior methods. We attribute this improvement to the robust correspondences established by DPCM. Although existing approaches obtain dense correspondences from accumulated point clouds captured by solid-state LiDARs, the accuracy of these correspondences is limited and does not generalize well to unseen scenarios. The substantial modality gap between LiDAR point clouds and camera images further hinders prior methods from achieving high pose estimation accuracy. Moreover, when applied to sparse point clouds captured by mechanical LiDARs, these methods suffer a severe degradation in performance, largely due to the pronounced differences in feature density between LiDAR and camera data. In contrast, our approach leverages cross-modal segmented masks to bridge the modality gap. The corner points extracted from these masks inherently encode reliable semantic information, which is critical for effective correspondence matching. DPCM not only increases the number of correspondences but also improves their geometric distribution, making them better suited for accurate pose estimation.

\subsection{Ablation Study and Analysis}
\label{sec.ablation_study}

\begin{table}[t!]
\caption{{Ablation study of structural and textural consistency on the 00 sequence of KITTI odometry (multi-scene optimization is not used). The best results are shown in bold type.}}
\centering
\settablefont
\begin{tabular}
{c@{\hspace{0.15cm}}c|cc|cc}
\toprule
\multicolumn{2}{c|}{\makecell{Consistency}} & \multicolumn{2}{c|}{Left Camera} & \multicolumn{2}{c}{Right Camera} \\
\hline
Structural & Textural & $e_r$ ($^\circ$)& $e_t$ (m) & $e_r$ ($^\circ$)& $e_t$ (m) \\
\hline
\hline

 &  & 2.222 & 0.792 & 2.600 & 0.896 \\
\checkmark &  & 1.227 & 0.304 & 1.533 & 0.411 \\
 & \checkmark & 2.416 & 0.827 & 2.633 & 0.984 \\
\checkmark & \checkmark & \textbf{1.125} & \textbf{0.278} & \textbf{1.425} & \textbf{0.354} \\

\bottomrule
\end{tabular}
\label{tab.ablation_consistency}
\end{table}

\begin{table*}[t!]
\caption{Ablation study of multi-view multi-scene joint optimization on the 00 sequence of KITTI odometry. The best results are shown in bold type.}
\centering
\fontsize{6.9}{10}\selectfont
\begin{tabular}{c@{\hspace{0.15cm}}c|c|c@{\hspace{0.15cm}}c|c@{\hspace{0.15cm}}c@{\hspace{0.15cm}}c|c@{\hspace{0.15cm}}c@{\hspace{0.15cm}}c|c@{\hspace{0.15cm}}c|c@{\hspace{0.15cm}}c@{\hspace{0.15cm}}c|c@{\hspace{0.15cm}}c@{\hspace{0.15cm}}c}
\toprule
\multicolumn{3}{c|}{Components}&\multicolumn{8}{c|}{Left Camera} &\multicolumn{8}{c}{Right Camera}\\
\hline
\multicolumn{2}{c|}{Multi-View}&\multirow{2}*{\makecell{Multi\\Scene}}&\multicolumn{2}{c|}{Magnitude}
&\multicolumn{3}{c|}{Rotation Error ($^\circ$)} &\multicolumn{3}{c|}{Translation Error (m)} 
&\multicolumn{2}{c|}{Magnitude}
&\multicolumn{3}{c|}{Rotation Error ($^\circ$)} &\multicolumn{3}{c}{Translation Error (m)}\\
\cline{1-2}
Intensity&Depth&&$e_r$ ($^\circ$) & $e_t$ (m) & Yaw & Pitch & Roll  & {X} &  {Y} &  {Z}   & $e_r$ ($^\circ$) & {$e_t$ (m)} & Yaw & Pitch & Roll   &  {X} &  {Y} &  {Z}\\
\hline
\hline
&& &1.625	&0.457	&0.820	&0.669	&0.899	&0.247	&0.232	&0.211
 &1.620	&0.472	&0.915	&0.691	&0.807	&0.257	&0.254	&0.197\\
\checkmark&&&1.387	&0.357	&0.755	&0.579	&0.711	&0.189	&0.205	&0.152
 &1.641	&0.459	&1.012	&0.679	&0.738	&0.257	&0.257	&0.178\\
\checkmark&\checkmark& &1.125	&0.278	&0.534	&0.534	&0.613	&0.134	&0.148	&0.136
&1.425	&0.354	&0.856	&0.563	&0.679	&0.186	&0.205	&0.138\\
&& \checkmark&0.406	&0.151	&0.180	&0.201	&0.223	&0.119	&0.058	&0.049
 &0.447	&0.192	&0.227	&0.243	&0.211	&0.148	&0.094	&0.051\\
\checkmark&& \checkmark&0.339	&0.106	&0.179	&\textbf{0.167}	&0.162	&0.069	&0.056	&0.039
&0.480	&0.138	&0.322	&0.239	&0.150	&0.096	&0.084	&0.033\\
\checkmark&\checkmark& \checkmark &\textbf{0.295}	&\textbf{0.078}	&\textbf{0.117}	&0.176	&\textbf{0.150}	&\textbf{0.051}	&\textbf{0.038}	&\textbf{0.032} &\textbf{0.336}	&\textbf{0.118}	&\textbf{0.216}	&\textbf{0.168}	&\textbf{0.121}	&\textbf{0.083}	&\textbf{0.067}	&\textbf{0.032}\\
\bottomrule
\end{tabular}
\label{tab.ablation_multi_vs_kitti}
\end{table*}

\begin{table}[t!]
\caption{Ablation study of multi-view multi-scene joint optimization on the 00 sequence of KITTI-360. The best results are shown in bold type.}
\centering
\fontsize{6.9}{10}\selectfont
\begin{tabular}{c@{\hspace{0.15cm}}c|c|c@{\hspace{0.15cm}}c|c@{\hspace{0.15cm}}c}
\toprule
\multicolumn{3}{c|}{Components}&\multicolumn{2}{c|}{Camera 1} &\multicolumn{2}{c}{Camera 2}\\
\hline
\multicolumn{2}{c|}{Multi-View}&\multirow{2}*{\makecell{Multi\\Scene}}&\multicolumn{2}{c|}{Magnitude}
&\multicolumn{2}{c}{Magnitude} \\

\cline{1-2}
Intensity&Depth&&$e_r$ ($^\circ$) & $e_t$ (m) & $e_r$ ($^\circ$) & {$e_t$ (m)}\\
\hline
\hline
&& &2.232	&0.524	&1.909	&0.440	\\
\checkmark&&&1.832	&0.402	&1.882	&0.415	\\
\checkmark&\checkmark& &1.496	&0.323	&1.506	&0.311	\\
&& \checkmark&0.914	&0.197	&0.686	&0.134	\\
\checkmark&& \checkmark&0.755	&0.135	&0.612	&0.085	\\
\checkmark&\checkmark& \checkmark &\textbf{0.605}	&\textbf{0.079}	&\textbf{0.498}	&\textbf{0.061}	\\
\bottomrule
\end{tabular}
\label{tab.ablation_multi_vs_kitti360}
\end{table}

To evaluate the algorithm's adaptability to incomplete and sparse point clouds, we further segmented the already limited field-of-view point clouds from the MIAS-LCEC-TF360 dataset. Specifically, as shown in Fig. \ref{fig.visualization_sparse_pcloud}, we divide each point cloud into six equal parts based on the recorded temporal sequence of the points. By progressively combining these segments, we create point clouds with varying densities, ranging from 1/6 (c1) to the full 6/6 (c6) density. 
The calibration results presented in Fig. \ref{fig.exp_on_different_density} show that EdO-LCEC achieves the smallest mean $e_r$ and $e_t$, along with the narrowest interquartile range, compared to other approaches across different point cloud densities. This demonstrates the stability and adaptability of EdO-LCEC under challenging conditions involving sparse or incomplete point clouds.

To validate the contribution of structural consistency and textural consistency in DPCM, we conducted an ablation study comparing their individual and combined effects. Multi-scene optimization is not used in this ablation study to better demonstrate the influence of the two consistencies. As shown in Table \ref{tab.ablation_consistency}, when both are used, the two components significantly improve calibration accuracy. Specifically, structural consistency preserves local geometric features through mask-guided alignment, while textural consistency evaluates visual similarity around matched correspondences. Their combination enables DPCM to maintain robust matching performance even in challenging scenarios with sparse or occluded point clouds,  which is a capability that existing methods (such as DVL and MIAS-LCEC) struggle to achieve. Additionally, MIAS-LCEC only compares the corner points within the matched segmented mask, while EdO-LCEC not only considers the corner points of the matched mask but also includes those of other unmatched masks. This allows EdO-LCEC to achieve global attention when searching for potential available matches when LiDAR frames are sparse.

Furthermore, we explore the contribution of the generalizable scene discriminator as well as the multi-view and multi-scene joint optimization. In this ablation study, the algorithm’s performance is comprehensively evaluated on the 00 sequence of KITTI odometry and KITTI-360, both with and without multi-view optimization (including both textural and spatial perspective views) and multi-scene optimization. The results presented in Table \ref{tab.ablation_multi_vs_kitti} and \ref{tab.ablation_multi_vs_kitti360} demonstrates that lacking any of these components significantly degrades calibration performance. In particular, calibration errors increase substantially when only single-view calibration is applied, as it lacks the comprehensive constraints provided by multi-view inputs. Additionally, the joint optimization from multiple scenes significantly improved the calibration accuracy compared to that under only multi-view optimization. These results confirm the advantage of incorporating spatial and textural constraints from multiple views and scenes, validating the robustness and adaptability of our environment-driven calibration strategy.

\subsection{Comparison of Correspondence Matching}
\label{sec.cmp_correspondence_matching}

\begin{figure}
    \centering
    \includegraphics[width=1.0\linewidth]{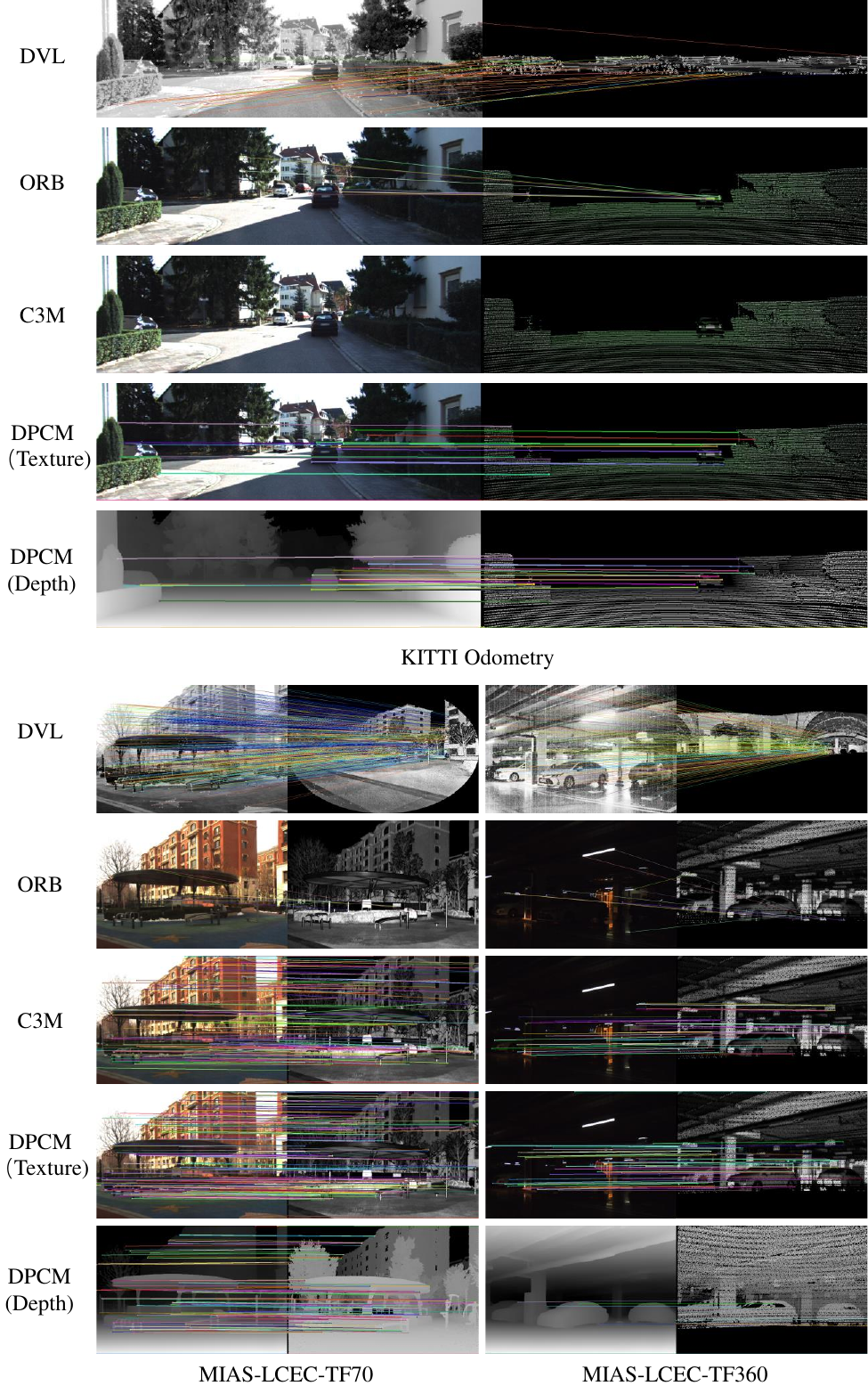}
    \caption{{Qualitative comparisons of correspondence matching.}}
    \label{fig.cmp_matching}
\end{figure}
{We present a comparison of correspondence matching results in Fig. \ref{fig.cmp_matching}. The visualized results indicate that our method} {significantly outperforms other correspondence matching algorithms employed in the comparison calibration methods. Specifically, our approach yields substantially more correct correspondences than ORB, DVL (which employs SuperGlue for direct 3D-2D correspondence matching), and C3M (the cross-modal mask matching algorithm provided by MIAS-LCEC). This demonstrates the superior adaptability of DPCM, particularly in scenarios involving sparse point clouds with minimal geometric and textural information.}

{The ORB algorithm relies on the fast Harris corner detector for keypoint extraction and employs a binary descriptor for efficient matching based on Hamming distance. While ORB performs well in image-to-image matching, it struggles with cross-modal feature matching, especially in LiDAR-camera integration scenarios. Conversely, while DVL and C3M perform adequately on dense point clouds such as those in the MIAS-LCEC-TF70, they face significant challenges when applied to the sparse point clouds in the KITTI odometry and MIAS-LCEC-TF360. The narrow overlapping field of view between LiDAR and camera in KITTI odometry creates difficulties in identifying reliable correspondences, particularly along the edges of the LiDAR field of view. This limitation significantly reduces the matching accuracy of DVL. Similarly, C3M, which restricts matching to corner points within aligned masks, struggles with sparse point clouds and limited sensor overlap.

Our proposed DPCM resolves these issues by incorporating both spatial and textural constraints. Unlike C3M, which treats mask instance matching as a strict constraint, DPCM uses it as a prior, enabling the generation of denser 3D-2D correspondences. This refinement significantly enhances the robustness of the algorithm in sparse environments and restricted fields of view, ensuring reliable calibration even under challenging conditions.}

\section{Conclusion}
\label{sec.conclusion}
In this article, we explore extending a new definition called ``environment-driven" for online LiDAR-camera extrinsic calibration. 
Unlike previous methods, our approach introduces environmental observation to maintain optimal performance across diverse and complex conditions.
Specifically, we designed a scene discriminator that can automatically observe the calibration scene. This discriminator can guide cross-modal feature matching by evaluating the feature density through multi-modal features extracted by large vision models. By leveraging structural and textural consistency between LiDAR projections and camera images, our method achieves more reliable 3D-2D correspondence matching. Additionally, we modeled the calibration process as a multi-view and multi-scene joint optimization problem, achieving high-precision and robust extrinsic matrix estimation through multi-view optimization within individual scenes and joint optimization across multiple scenarios. Extensive experiments on real-world datasets demonstrate that our environment-driven calibration strategy achieves the state-of-the-art performance.

\clearpage
\clearpage
\setcounter{page}{1}

{
   \newpage
       \twocolumn[
        \centering
        \Large
        \textbf{Environment-Driven Online LiDAR-Camera Extrinsic Calibration}\\
        \vspace{0.5em}Supplementary Material \\
        \vspace{1.0em}
       ]
}

\section{Details of the Algorithm Implementation}
\label{sec.sup_algo_detail}
\begin{figure*}
    \centering
    \includegraphics[width=1\linewidth]{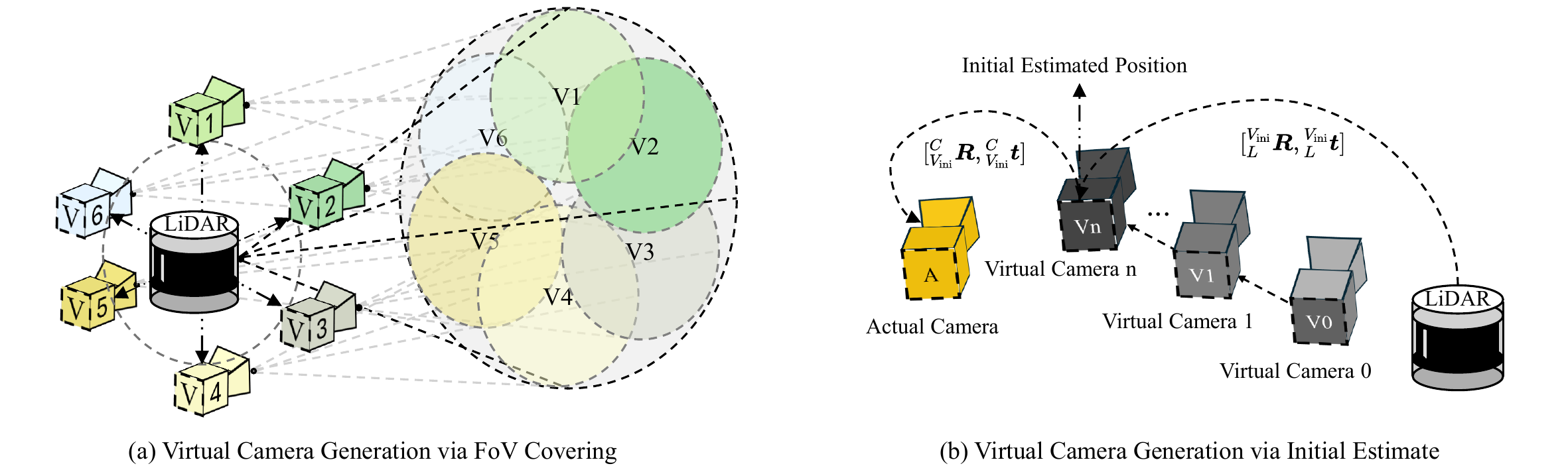}
    \caption{Possible automatic virtual camera generation method: (a) generating virtual cameras by covering the range of the entire LiDAR scan; (b) generating virtual cameras by distributing them along the translation vector of the initial guess.}
    \label{fig.virtual_cam_gen_alternative}
\end{figure*}

\subsection{Estimation of the 4-DoF Similarity Transformation}
As presented in the ``Dual-Path Correspondence Matching'' in Sect. \ref{sec.dualpath_matching}, we initially adopted the method described in \cite{zhiwei2024lcec} for cross-modal mask matching. \cite{zhiwei2024lcec} proved that when two 3D LiDAR points $\boldsymbol{q}^V$ (reference) and $\boldsymbol{p}^V$ (target) in the virtual camera coordinate system are close in depth, the relationship of their projection coordinates $\tilde{\boldsymbol{p}}_{v,c}$ and $\tilde{\boldsymbol{q}}_{v,c}$ on virtual image and camera image can be expressed by a following 4-DoF similarity transformation:
\begin{equation}
\begin{pmatrix}
\tilde{\boldsymbol{p}}_c & \tilde{\boldsymbol{q}}_c \\
\end{pmatrix}
=
\begin{pmatrix}
s{\boldsymbol{R}}^A & \boldsymbol{t}^A \\
\boldsymbol{0}^\top &1 \\
\end{pmatrix}
\begin{pmatrix}
\tilde{\boldsymbol{p}}_v & \tilde{\boldsymbol{q}}_v \\
\end{pmatrix}
\label{eq.LscLaw_final}
\end{equation}
where $\boldsymbol{R}^A\in{SO(2)}$ represents the rotation matrix, $\boldsymbol{t}^A$ denotes the translation vector, and $s$ represents the scaling factor. We can assume that after the similarity transformation, any points within a given mask in the LIP image perfectly align with the corresponding points from the RGB image, and thus:
\begin{equation}
\begin{aligned}
\hat{\boldsymbol{c}}^V = s\boldsymbol{R}^A \boldsymbol{c}^V + \boldsymbol{t}^A, \\
\hat{\boldsymbol{o}}^V = s\boldsymbol{R}^A \boldsymbol{o}^V + \boldsymbol{t}^A. 
\end{aligned}
\label{eq.correction}
\end{equation}
where $\hat{\boldsymbol{o}}^V$ and $\hat{\boldsymbol{c}}^V$ represent the updated coordinates of the instance's center and corner points in the virtual image, respectively. In DPCM, we adopt the method proposed in \cite{zhiwei2024lcec} to estimate the similarity transformation, which serves as a semantic prior to guide dense correspondence matching between the LiDAR and camera modalities. When the calibration scenarios become complex or the initial misalignment between the LiDAR and camera is relatively large, the estimation of the rotational-scale component $s\boldsymbol{R}^A$ in the similarity transformation may become unstable. In such real-world cases, $s\boldsymbol{R}^A$ can be approximated by an identity matrix, allowing the transformation process to primarily depend on the translation component $\boldsymbol{t}^A$.

\subsection{Components in the Structural Consistency}
In the algorithm implementation, the structural consistency component $H(\hat{\boldsymbol{e}}^V_{i,k},\boldsymbol{e}^C_{j,k},\boldsymbol{e}^C_{j,K-k+1})$ introduced in Sect. \ref{sec.dualpath_matching} can be defined as follows:
\begin{equation}
\begin{aligned}
H(\hat{\boldsymbol{e}}^V_{i,k},\boldsymbol{e}^C_{j,k},{\boldsymbol{e}}^C_{j,K-k+1}) = 
\min\{
&S(\hat{\boldsymbol{e}}^V_{i,k},\boldsymbol{e}^C_{j,k}),\\
&S(\hat{\boldsymbol{e}}^V_{i,k},\boldsymbol{e}^C_{j,K-k+1})\}.
\label{eq._struct_similarity}
\end{aligned}
\end{equation}
The function $S$ in \eqref{eq._struct_similarity} measures the similarity between a neighboring vertex of the current corner point and that of the corresponding target corner point, and is formulated as:
\begin{equation}
S(\hat{\boldsymbol{e}}^V_{i,a},\boldsymbol{e}^C_{j,b}) =w
{\frac{\| (\hat{\boldsymbol{e}}^V_{i,a} - \hat{\boldsymbol{c}}^V_{i} ) -(\boldsymbol{e}^C_{j,b} - \boldsymbol{c}_j^C )\|_2}{\max{\{\|\hat{\boldsymbol{e}}^V_{i,a} - \hat{\boldsymbol{c}}^V_{i}  \|_2, \| \boldsymbol{e}^C_{j,b} - \boldsymbol{c}_j^C \|_2}\}}},
\label{eq._struct_similarity_s}
\end{equation}
where $w$ is a constant scale parameter.

\subsection{Components in the Multi-View Optimization}
For the multi-view optimization, as presented in Sect. \ref{sec.st_optimization}, we utilize a Gaussian kernel $\mathcal{K}(d'_{j},\bar{d'})$ to calculate the depth-normalized reprojection error. In detail, $\mathcal{K}(d'_{j},\bar{d'})$ is defined as follows:
\begin{equation}
\mathcal{K}(d'_{j},\bar{d'}) = \exp{ \bigg(\frac{ -\left\|{d'_{j}} - \bar{d'}\right\|^2 }{{2\bar{d_g}}^2} \bigg)},
\end{equation}
where $\bar{d_g}$ is the true average depth of the correspondences. The normalized depth ${d'_j}$ in the equation is the normalization result of the point cloud depth in the range $[0,1]$. The calculation of ${d'_j}$ is defined as follows:
\begin{equation}
{d'_j} = \frac{d_j - d_{\min}}{d_{\max} - d_{\min}},
\end{equation}
where $d_{\min}$ and $d_{\max}$ are the minimum and maximum depth values of the input point cloud in the virtual camera perspective.

In our algorithm implementation, we use the pinhole model as the projection model to calculate the reprojection error presented in (\ref{eq.multi_view}) and (\ref{eq.multi_scene}). When the camera intrinsic matrix $\boldsymbol{K}$ is known, the 3D LiDAR point $\boldsymbol{p}^{L}=[x^L,y^L,z^L]^\top$ can be projected onto a 2D image pixel $\boldsymbol{{p}} = [u,v]^\top$ using the following expression:
\begin{equation}
\pi(\boldsymbol{p}^{L}) = \tilde{\boldsymbol{p}} = \frac{\boldsymbol{K}\boldsymbol{p}^{L}}{(\boldsymbol{p}^{L})^\top\boldsymbol{1}_{z}},
\end{equation}
where $\tilde{\boldsymbol{p}}$ represents the homogeneous coordinates of $\boldsymbol{{p}}$ and $\boldsymbol{1}_{z}=[0,0,1]^\top$. Given the extrinsic transformation ${^C_L}\boldsymbol{T}$, the reprojection between the correspondence $(\boldsymbol{p}_{i}^L, \boldsymbol{p}_{i})$ can be obtained as follows:
\begin{equation}
\epsilon_i = \left\|\pi( {^C_L}\boldsymbol{T}\boldsymbol{p}_i^{L}) - \tilde{\boldsymbol{p}}_i  \right\|_2= \left\|\frac{\boldsymbol{K}({^{C}_{L}\boldsymbol{R}}\boldsymbol{p}_i^{L} + {^{C}_{L}\boldsymbol{t}})}{({^{C}_{L}\boldsymbol{R}}\boldsymbol{p}_i^{L} + {^{C}_{L}\boldsymbol{t}})^\top\boldsymbol{1}_{z}} - \tilde{\boldsymbol{p}}_i\right\|_2.
\end{equation}
By applying the Gaussian normalization $G$ given in (\ref{eq.dpcm_gaussian}), the reprojection errors of different correspondences are provided with weights based on the depth distribution.

\subsection{Configuration of Multi-Scene Optimization}

As shown in Table \ref{tab.ablation_multi_vs_kitti} and \ref{tab.ablation_multi_vs_kitti360} in Sect. \ref{sec.ablation_study}, multi-scene optimization significantly improves calibration accuracy compared to multi-view optimization. Using more frames leads to greater accuracy improvements from multi-scene optimization. In our experiments, we used 20 frames to obtain the results reported in the table. For real-world applications, we recommend using at least 5 frames to achieve stable multi-scene calibration. When the point cloud is relatively dense, as in the TF70 dataset, a single frame is enough to produce accurate results by using the multi-view optimization.

\subsection{Discussion on Virtual Camera Generation}

As a crucial component of the scene discriminator, the virtual camera generation process ensures proper positioning for capturing LiDAR projections. The primary goal of virtual camera generation is to maximize the capture of matchable features for correspondences between LiDAR point clouds and camera images. Therefore, virtual cameras must be judiciously positioned and deployed in sufficient numbers. EdO-LCEC estimates the feature density of the environment, derived from LiDAR projections and camera images, to determine the number of virtual cameras required to balance calibration coverage and computational cost. The virtual cameras are arranged along the $X$, $Y$, and $Z$ axes on a sphere centered at the LiDAR perspective origin. This adaptive generation approach allows dynamic adjustment of the virtual camera count according to environmental complexity, making it suitable for diverse and uncertain operating conditions. For instance, in the relatively sparse KITTI dataset, more virtual cameras (around 7) may be needed to ensure adequate coverage, whereas in dense scenarios such as the TF70 dataset, fewer virtual cameras (about 1-2) suffice, avoiding unnecessary computational overhead. Extensive experiments in Sect. \ref{sec.experiment} demonstrates that this feature-density-based generation strategy performs robustly across challenging scenarios and exhibits strong environmental adaptivity. We contend that this virtual camera generation method is optimal within our proposed environment-driven calibration framework.

\begin{table*}[t!]
\caption{Details of comparison methods.}
\centering
\fontsize{6.9}{8}\selectfont
\begin{tabular}{lrccccc}
\toprule
Methods & Publication &Type &Parameterization or Basic Tools &Computation Speed &Accuracy &Provide Code? \\
\hline
UMich \cite{pandey2015automatic} &JFR 2015&MI-based& Mutual Information &  Medium & Medium & Yes\\
RegNet \cite{schneider2017regnet} &IV 2017&Learning-based& CNN, Direct Pose Regression &  Medium & Medium & No\\
CalibNet  \cite{iyer2018calibnet} &IROS 2018&Learning-based& Geometrically Supervised
DNN & High  & Low & Yes\\
RGGNet \cite{yuan2020rggnet} &RAL 2020&Learning-based & Deep Learning Network, Riemannian
Geometry &  High & Medium & Yes\\
CalibRCNN \cite{shi2020calibrcnn}&IROS 2020 &Learning-based& DCNN, LSTM, Direct Pose Regression & Medium &  Medium & No\\
CalibDNN \cite{zhao2021calibdnn} &SPIE 2021&Learning-based& CNN, Direct Pose Regression &  Medium & Medium & No\\
LCCNet \cite{lv2021lccnet} &CVPR 2021&Learning-based& CNN, Cost Volume, Direct Pose Regression & High  & Medium & Yes \\
HKU-Mars  \cite{yuan2021pixel} &RAL 2021&Edge-based& Adaptive Voxel & Low & Very Low & Yes\\
CRLF \cite{ma2021crlf} &arXiv 2021 &Semantic-based&Segmented Road Pole and Lane, Perspective-3-Lines &  Low & Low & Yes\\
DVL  \cite{koide2023general} &ICRA 2023&Point-based& SuperGlue, RANSAC, Normalized Information Distance &  Medium & High & Yes\\
Borer \etal \cite{borer2024chaos} &WACV 2024&MI-based& Geometric Mutual Information  & High & High  & No \\
MIAS-LCEC \cite{zhiwei2024lcec} &T-IV 2024 &Semantic-based& MobileSAM, C3M, PnP &  Low & High & Yes\\

\bottomrule
\end{tabular}
\label{tab.cmp_methods}
\end{table*}
In our preliminary experiments, we attempted to fix the number of virtual cameras according to the complexity of the calibration data. In relatively simple environments with few dynamic objects, such as a controlled calibration room, this method proved stable and easy to configure, given available computational resources. While this manual strategy enhanced calibration performance in certain specific cases, it lacked generalizability across more complex and diverse scenarios. The required parameters vary with environmental changes, and the absence of environmental perception prevents the algorithm from automatically adapting to unseen and challenging conditions.

Apart from this manual way to generate virtual cameras, we have also considered two possible alternatives that can achieve automatic generation. Here, we present these strategies and provide tailored application recommendations for different scenarios:

\begin{enumerate}

  \item[(a)]  
  One of the automatic methods to determine the number of virtual cameras is to compute the ratio of LiDAR to camera fields of view, which allows deducing the minimum number of virtual cameras required. As depicted in Fig. \ref{fig.virtual_cam_gen_alternative}. (a), each virtual camera matches the intrinsic FoV of the real camera, and collectively they cover the entire LiDAR scan range. A straightforward formula for determining the required number of virtual cameras is:

\begin{equation}
n = \rho \left\lceil \frac{\mathrm{FoV}_{\mathrm{LiDAR}}}{\mathrm{FoV}_{\mathrm{cam}}} \right\rceil,
\end{equation}

where \(\mathrm{FoV}_{\mathrm{LiDAR}}\) is the LiDAR’s field of view, \(\mathrm{FoV}_{\mathrm{cam}}\) is the camera’s field of view, and \(\lceil \cdot \rceil\) denotes the ceiling function to ensure full coverage of the LiDAR scan. The parameter \(\rho\) defines the virtual camera density when the FoV ratio between the LiDAR and camera is equal to 1.
If the FoVs of LiDAR and camera are unknown. We can estimate them using the following method: \\

For the FoV of LiDAR, we can obtain an estimate of the FoV using the captured point clouds. Let the LiDAR point cloud be $\mathcal{P}=\{\boldsymbol{p}_i = [x_i,y_i,z_i]^\top\}_{i=1}^N$ in the sensor frame. For each point $\boldsymbol{p}_i=[x_i,y_i,z_i]^\top$ define the horizontal angle $\theta_i$ and vertical angle $\phi_i$ by the following equation:
\begin{equation}
\theta_i = \operatorname{atan2}(y_i, x_i),\quad
\phi_i = \operatorname{atan2}\bigl(z_i, \sqrt{x_i^2+y_i^2}\bigr).
\end{equation}
The horizontal and vertical fields of view can then be obtained as follows:
\begin{equation}
\begin{aligned}
\mathrm{FoV}_{\mathrm{h}, \mathrm{LiDAR}} = \max_i \theta_i - \min_i \theta_i,
\\
\mathrm{FoV}_{\mathrm{v},\mathrm{LiDAR}} = \max_i \phi_i - \min_i \phi_i.
\end{aligned}
\end{equation}
For the camera, its FoV can be computed as follows:
\begin{equation}
\mathrm{FoV}_{\text{cam}} = 2\arctan\!\left(\frac{s}{2f}\right),
\end{equation}
where $s$ is the relevant sensor dimension (e.g., width or height) and $f$ is the focal length. Using these quantities, one can decide virtual camera counts via:
\begin{equation}
n = \rho \left\lceil \frac{\mathrm{FoV}_{\mathrm{h},\mathrm{LiDAR}}}{\mathrm{FoV}_{\mathrm{h},\mathrm{cam}}} \right\rceil \quad \mathrm{or} \quad n = \rho \left\lceil \frac{\mathrm{FoV}_{\mathrm{v},\mathrm{LiDAR}}}{\mathrm{FoV}_{\mathrm{v},\mathrm{cam}}} \right\rceil
\end{equation}
with the choice of horizontal or vertical FoV depending on the coverage requirement. This FoV-based approach is particularly effective when the FoV difference between the LiDAR and camera is moderate. If the FoV gap is too large, a substantial portion of the projection may remain unused for feature matching. Conversely, if the FoV gap is too small, too few projections are produced, which can cause failures in scenarios with sparse LiDAR point clouds. While this strategy overcomes the feature matching difficulties introduced by FoV discrepancies, it does not address the modality gap or the differences in feature density between LiDAR and camera. Moreover, the absence of an environmental perception mechanism limits its adaptability, making it unsuitable for an environment-driven calibration framework.

  \item[(b)] 
  If an initial estimate of the extrinsic parameters is available, the number of virtual cameras can also be determined by the distance between the LiDAR origin and the initial estimate. As illustrated in Fig. \ref{fig.virtual_cam_gen_alternative}. (b), given $m$ virtual cameras per unit distance, the number of virtual cameras $n$ can be calculated as
\begin{equation}
n = m\left\|\boldsymbol{t}_{\text{ini}}\right\|_2,
\end{equation}
and their positions are distributed along the translation vector from the LiDAR origin to the initial estimate. This generation strategy is guided by prior calibration results, making it suitable for relatively fixed sensor configurations with moderate extrinsic variation. By positioning virtual cameras along the vector from the LiDAR origin to the initial estimated camera position, this approach can increase the overlapping field of view. However, it does not account for modality differences or feature density. In scenarios where LiDAR point clouds are considerably sparser than camera images, such as in KITTI and nuScenes, this method cannot effectively determine the optimal number of virtual cameras to balance calibration accuracy and computational efficiency. Furthermore, in robotic applications such as multi-robot joint perception, obtaining an initial estimate between sensors on different platforms is challenging. The method also depends on the accuracy of the initial guess; if the estimate is unavailable or inaccurate, the approach becomes ineffective. Given these limitations and the lack of an environmental perception mechanism, this strategy is impractical for environment-driven calibration.
\end{enumerate}

We believe that these alternative methods for generating virtual cameras can be applied across a range of real-world scenarios. However, rather than seeking an elusive “best” method, it is more important to choose the most suitable configuration based on the specific application. We believe that concerning an algorithm module in an environment-driven framework, there is ``no free lunch". Users may balance the computational cost and calibration accuracy, selecting the approach that best suits their specific application to achieve more stable results.  If the users are not sure which strategy to use, we strongly recommend using the default feature-density-based method, which provides a stable and generalizable approach for generating a sufficient number of virtual cameras.

\section{Calibration Parameters}
To foster future research toward LiDAR-camera sensor
fusion, we release the calibration parameters that we obtained with our method for the 00 sequence of the KITTI odometry. In detail, we provide the rotation matrices and translation vectors for projecting the LiDAR point cloud into the camera image or rendering LiDAR point clouds with colors. 

\noindent \textbf{Calibration Parameters for Left RGB Camera}
\[
\begin{aligned}
\boldsymbol{R} &= 
\begin{bmatrix}
-2.5863e-04 &-9.9997e-01 &-7.5239e-03 \\
-6.8893e-03 &7.5255e-03 &-9.9995e-01 \\
9.9998e-01 &-2.0678e-04 &-6.8911e-03 \\
\end{bmatrix} \\
\boldsymbol{t} &= 
\begin{bmatrix}
0.070478 &-0.057913 &-0.286353
\end{bmatrix}^\top
\end{aligned}
\]

\noindent \textbf{Calibration Parameters for Right RGB Camera}
\[
\begin{aligned}
\boldsymbol{R} &= 
\begin{bmatrix}
-5.3770e-04 &-9.9998e-01 &-6.6640e-03 \\
-8.1238e-03 &6.6681e-03 &-9.9994e-01 \\
 9.9997e-01 &-4.8353e-04 &-8.1272e-03 \\
\end{bmatrix} \\
\boldsymbol{t} &= 
\begin{bmatrix}
-0.494557 &-0.042731 &-0.261487 
\end{bmatrix}^\top
\end{aligned}
\]

Examples of the data fusion result using our extrinsic calibration parameters in KITTI odometry are shown in the demo videos on our website.

\section{Details of Experimental Configuration}
\subsection{Details of Comparison Methods}

The details of the comparison methods are illustrated in Table \ref{tab.cmp_methods}. For algorithms that did not release their code, we directly reference the results published in their papers. Notably, RegNet \cite{schneider2017regnet}, Borer \etal \cite{borer2024chaos}, and CalibDNN \cite{zhao2021calibdnn} employ different experimental datasets and evaluation processes, which limits the potential for a fully quantitative comparison. Furthermore, these methods do not specify which frames from the KITTI dataset were used for training and testing. Therefore, we rely on the published results to compare these approaches to others on the KITTI odometry 00 sequence.

\begin{table}[t!]
\caption{Dataset statistics of KITTI odometry 00-09 sequences.}
\centering
\fontsize{6.9}{10}\selectfont
\begin{tabular}{c|@{\hspace{0.2cm}}c@{\hspace{0.2cm}}c@{\hspace{0.2cm}}c@{\hspace{0.2cm}}c@{\hspace{0.2cm}}c@{\hspace{0.2cm}}c@{\hspace{0.2cm}}c@{\hspace{0.2cm}}c@{\hspace{0.2cm}}c@{\hspace{0.2cm}}c}
\toprule
& 00 &01 &02 &03 &04 &05 &06 &07 &08 &09 \\
\hline
Initial Pairs & 4541 &1101& 4661 & 801 & 271 & 2761 &1101 & 1101 &4071 &4091\\
Downsampling & 10 &10& 10 & 10 & 2 & 10 &10 & 10 &10 &10\\
Evaluated Pairs & 455 &111& 467 & 81 & 136 & 277 &111 & 111 &408 &410\\
\bottomrule
\end{tabular}
\label{tab.kitti_dataset_statistics}
\end{table}

\begin{figure*}[t!]
    \centering
    \includegraphics[width=0.99\linewidth]{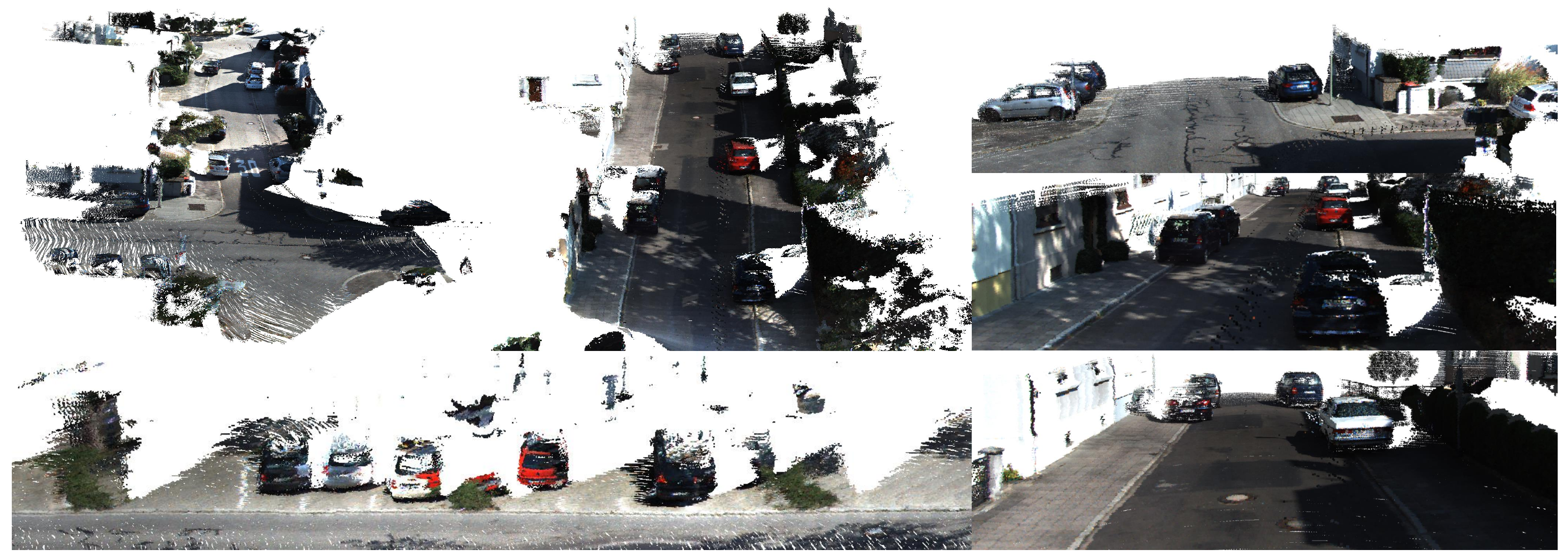}
    \caption{An example of the data fusion result on KITTI odometry using our calibration parameters.}
    \label{fig.exp_detail_datafusion_on_KITTI}
\end{figure*}

\begin{figure*}[t!]
    \centering
    \includegraphics[width=1\linewidth]{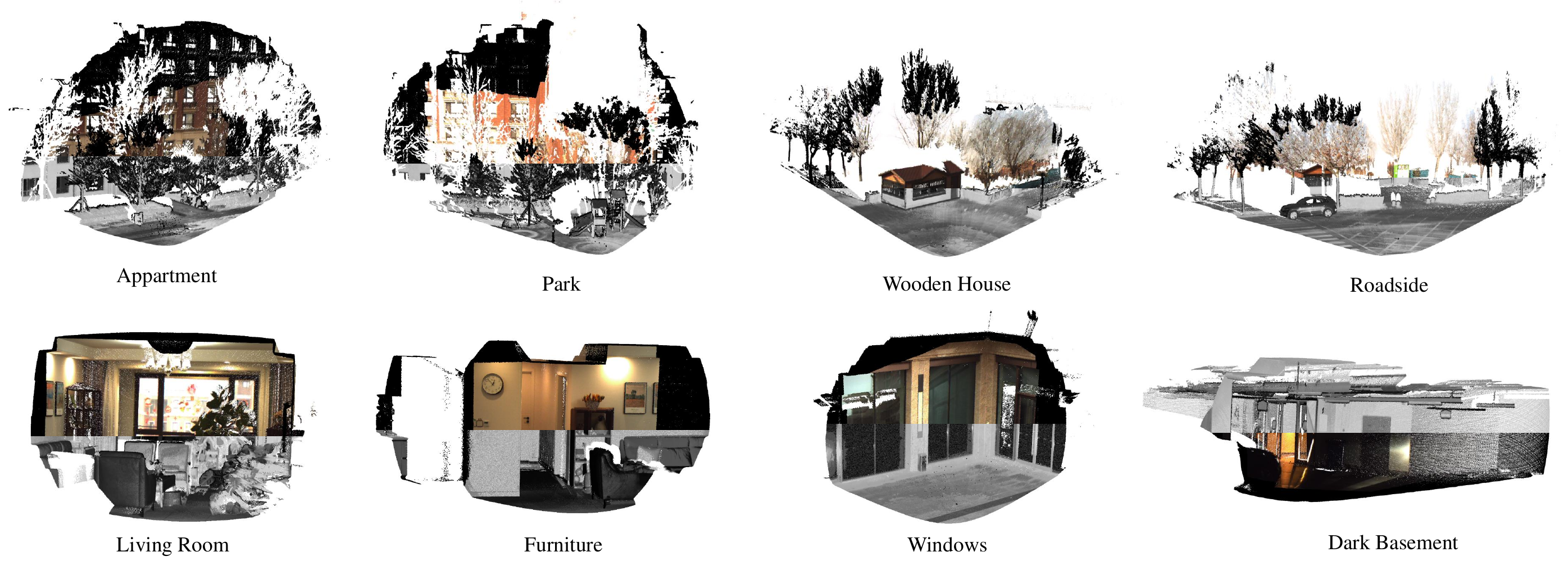}
    \caption{Visualization of EdO-LCEC calibration results through LiDAR and camera data fusion on MIAS-LCEC datasets.}
    \label{fig.EdO_calib_visualization}
\end{figure*}

\subsection{Evaluation Metrics}
To comprehensively evaluate the performance of LCEC approaches, we follow the previous works \cite{lv2021lccnet,zhiwei2024lcec,iyer2018calibnet,koide2023general}, using the magnitude $e_r$ of Euler angle error and the magnitude $e_t$ of the translation error, with the following expression:
\begin{equation}
\begin{aligned}
e_r &=  \left\|{{^C_L}\boldsymbol{r}^*} - {{^C_L}\boldsymbol{r}}\right\|_2, \\
e_t &= \left\|-{({^{C}_{L}\boldsymbol{R}^*})^{-1}}{{^C_L}\boldsymbol{t}}^* +{{^{C}_{L}\boldsymbol{R}}^{-1}}{{^C_L}\boldsymbol{t}}\right\|_2,
\end{aligned}
\label{eq.exp_metrics}
\end{equation}
{to quantify the calibration errors. In (\ref{eq.exp_metrics}), ${{^C_L}\boldsymbol{r}^{*}}$ and ${{^C_L}\boldsymbol{r}}$ represent the estimated and ground-truth Euler angle vectors, computed from the rotation matrices ${^{C}_{L}\boldsymbol{R}}^*$ and ${^{C}_{L}\boldsymbol{R}}$, respectively. Similarly, ${^C_L}\boldsymbol{t}^*$ and ${^C_L}\boldsymbol{t}$ denote the estimated and ground-truth translation vectors from LiDAR to camera, respectively.} $-{{^{C}_{L}\boldsymbol{R}}^{-1}}{{^C_L}\boldsymbol{t}}$ is the translation from LiDAR pose to camera pose when (\ref{eq.lidar_to_camera_point}) is used to depict the point translation.

\subsection{Data Preparation}

We have conducted extensive experiments on KITTI odometry, KITTI360, nuScenes, MIAS-LCEC-TF70, and MIAS-LCEC-TF360. Now we provide our data preparation details.
\begin{itemize}
\item{\textit{\textbf{KITTI odometry}}:
KITTI odometry is a large-scale public dataset recorded using a vehicle equipped with two RGB cameras Point Grey Flea 2 and one LiDAR of type Velodyne HDL-64E. Sensor data is captured at 10 Hz. We utilize the first 10 sequences (00-09) for evaluation. The dataset statistics of KITTI odometry are presented in Table \ref{tab.kitti_dataset_statistics}. It can be observed that there are a total of 24500 pairs of point clouds and images in KITTI odometry. Note that most of the SoTA comparison methods are slow to complete calibration with a single frame. For the comparison algorithms that provided source code, we performed calibration every ten frames (except for 2 frames in the 04 sequence), enhancing the efficiency of the experiments without compromising the validity of the evaluation.}

\item{\textit{\textbf{KITTI-360}}:
The KITTI-360 dataset is a large-scale autonomous driving dataset designed to advance research in various areas such as semantic scene understanding, 3D reconstruction, and motion forecasting. We utilize the images captured from the front stereo cameras and the point clouds scanned by the Velodyne HDL-64E LiDAR. We use the first 3000 pairs of LiDAR point clouds and camera images in the 00 sequence to test the performance of each component of EdO-LCEC. Similarly, we sampled the raw datasets every 10 frames.}
\item{\textit{\textbf{nuScenes}}:
The LiDAR sensor used in nuScenes is a 32-line spinning LiDAR (Velodyne HDL32E), which is considerably sparser than the 64-line LiDARs employed in KITTI and KITTI-360, thereby making the calibration task more challenging. To the best of our knowledge, few calibration methods have been evaluated on nuScenes. Following the official dataset split, we employ the nuScenes SDK to generate image-point cloud pairs from the 150 testing scenes in the v1.0-test set to construct our evaluation data. Since each scene in nuScenes contains only a limited number of sensor samples, we sample one out of every ten scenes to ensure the relative completeness of the data sequence within each scene.
}

\item{\textit{\textbf{MIAS-LCEC-TF70}}:
MIAS-LCEC-TF70 is a diverse and challenging dataset that contains 60 pairs of 4D point clouds (including spatial coordinates with intensity data) and RGB images, collected using a Livox Mid-70 LiDAR and a MindVision SUA202GC camera, from a variety of challenging indoor and outdoor environments. We compare our method with SoTA approaches on the total six subsets to comprehensively evaluate the algorithm's performance.}
\item{\textit{\textbf{MIAS-LCEC-TF360}}:
MIAS-LCEC-TF360 contains 12 pairs of 4D point clouds and RGB images, collected using a Livox Mid-360 LiDAR and a MindVision SUA202GC camera from both indoor and outdoor environments. Livox Mid-360 LiDAR produces a sparser point cloud compared to that generated by the Livox Mid-70 LiDAR. The significant difference in the FoV between this type of LiDAR and the camera results in only a small overlap in the collected data. Consequently, this dataset is particularly well-suited for evaluating the adaptability of algorithms to challenging scenarios.}
\end{itemize}  

Notably, an initial guess is recommended if the LiDAR and camera have a large difference in position or direction. For example, in the nuScenes dataset, the front camera and the LiDAR are installed in totally different directions with no shared field of view. We configure an initial offset on the translation and rotation to ensure that the initial mis-calibration rotation error is within 2 Euler degrees along each axis and the total translation error is within 0.5m.

\section{More Visualization Results}
\label{sec.more_visualization}

In this section, we present detailed visualization results of EdO-LCEC through LiDAR-camera data fusion on the KITTI Odometry and MIAS-LCEC datasets. An example of the fusion result using our calibrated extrinsic parameters for the left camera in KITTI odometry is illustrated in Fig. \ref{fig.exp_detail_datafusion_on_KITTI}. The LiDAR point clouds and camera RGB images are precisely aligned under the estimated calibration parameters, producing highly consistent geometric and photometric correspondences. As shown in Fig. \ref{fig.EdO_calib_visualization}, the LiDAR projections and camera images exhibit near-perfect alignment along both textural and structural edges, confirming the sub-pixel accuracy achieved by our method. Furthermore, our algorithm maintains high robustness under challenging conditions, including poor illumination, adverse weather, and scenes with sparse or ambiguous geometry. The ability to preserve geometric consistency across such conditions highlights the strong generalization capacity of our environment-driven strategy. Beyond quantitative metrics, these visualization results intuitively demonstrate the practical impact of EdO-LCEC. By enabling seamless fusion of 3D spatial geometry and 2D visual texture, it allows robots to perceive, localize, and interact with their surroundings in a manner closer to human visual understanding. This perceptual synergy between LiDAR and camera further paves the way toward adaptive, context-aware robotic systems capable of reliable operation in complex real-world environments.

\end{document}